\documentclass[journal]{IEEEtran}
\IEEEoverridecommandlockouts
\usepackage{cite} 
\usepackage{amsmath,amssymb,amsfonts}
\usepackage{graphicx}
\usepackage{textcomp}
\usepackage{multirow}
\usepackage{algorithm}
\usepackage{algorithmic}
\usepackage{pifont}
\newcommand{\cmark}{\ding{51}}%
\newcommand{\xmark}{\ding{55}}%
\usepackage[dvipsnames]{xcolor}
\usepackage{colortbl}   %
\usepackage{booktabs}   %
\usepackage{adjustbox}
\usepackage{tikz}
\usepackage{dblfloatfix}
\usepackage{fancyhdr}
\usepackage{array}
\usepackage{xspace}

\usepackage{subfigure}
\usepackage{hyperref}

\def\BibTeX{{\rm B\kern-.05em{\sc i\kern-.025em b}\kern-.08em
    T\kern-.1667em\lower.7ex\hbox{E}\kern-.125emX}}

\newcommand{\bstars}{\textbf{\textsuperscript{$\bigstar$}}}

\providecolor{wsLine}{RGB}{0,0,0}
\providecolor{wsHatch}{RGB}{31,78,121}
\providecolor{wsHigh}{RGB}{33,113,181}
\providecolor{wsMid}{RGB}{239,243,255}
\providecolor{wsLow}{RGB}{107,174,214}
\usetikzlibrary{patterns}

\providecommand{\weshare}{\textsc{BitShare}}
\providecommand{\weshareone}{\textsc{BitShare-U}}
\providecommand{\wesharetwo}{\textsc{BitShare-N}}

\begin{document}

\title{BitShare: Sensitivity-Aware Binning and Bit Sharing \\for Compacting LLM Weight Storage}

\author{Qiuyu Ren, Sudipta Paria, Aritra Dasgupta, Swarup Bhunia\\
Department of Electrical and Computer Engineering (ECE), University of Florida, Gainesville, FL 32611, USA \\
\{qiuyuren,sudiptaparia,aritradasgupta\}@ufl.edu, swarup@ece.ufl.edu}
\maketitle

\begin{abstract}
Decoding with large language models at small batch sizes is limited by memory bandwidth, so the bits stored per
weight largely set its cost. 
Group-wise integer quantization stores every weight of a group at the same precision,
although the input channels of a group differ in sensitivity. 
We present \weshare{}, a codebook-free weight format that keeps the integer
grid and the dequantizer of group-wise quantization but varies precision within each group. Sensitivity-aware binning
places the channels of each group into three bins at fixed positions, and static bit sharing with optimal bit-value allocation lets the less sensitive
weights share their lower bits, each stored once per group or once per pair of weights. A sharing-aware version of
GPTQ chooses every stored value, with no fine-tuning and no codebook. \weshare{} has two modes. \weshareone{} gives every
group the same layout, so its decoder routes bits by fixed wiring, while \wesharetwo{} chooses a layout per group within
the same bit budget. With calibration on the Pile and at the same storage, both modes have lower perplexity than GPTQ at 3 and 2 bits on WikiText-2 and C4. At 2 bits across Llama-2 and Llama-3 models, \wesharetwo{} lowers WikiText-2 perplexity by up to 70\% against SliM-LLM
and up to 35\% against GPTAQ. At the 2-bit perplexity of SliM-LLM, \weshareone{} stores 13.6-19.4\% fewer bits per weight, cutting DRAM weight traffic by the same fraction. A verified hardware decoder implementation synthesized in SkyWater 130\,nm reconstructs \weshare{} weights at the throughput of INT-$n$ and SliM-LLM decoders.
\end{abstract}

\section{Introduction}
\label{sec:intro}

\usetikzlibrary{patterns}

\definecolor{wsLine}{RGB}{0,0,0}         %
\definecolor{wsHatch}{RGB}{31,78,121}
\definecolor{wsCode}{RGB}{158,202,225}   %
\definecolor{wsHigh}{RGB}{33,113,181}    %
\definecolor{wsMid}{RGB}{239,243,255}    %
\definecolor{wsLow}{RGB}{107,174,214}    %
\definecolor{wsOut}{RGB}{8,48,107}       %

\begin{figure*}[t]
\centering
\begin{tikzpicture}[x=0.15mm,y=2.1mm,font=\scriptsize,
  code/.style={fill=wsCode,draw=wsLine},
  hi/.style={fill=wsHigh,draw=wsLine},
  mid/.style={fill=wsMid,draw=wsLine},
  lo/.style={fill=wsLow,draw=wsLine},
  shf/.style={pattern=north east lines,pattern color=wsHatch,draw=wsLine},
  shc/.style={pattern=crosshatch dots,pattern color=wsHatch,draw=wsLine},
  rate/.style={densely dashed,line width=0.9pt,wsOut},
  name/.style={anchor=east,align=right,text width=31mm},
  info/.style={anchor=west,align=left,text width=47mm}]

  \def\gx#1{#1*140}
  \def\xend{548}

  \def\b{0}
  \foreach \g in {0,1,2,3} \draw[code] (\g*140,\b) rectangle ++(128,2);
  \draw[rate] (-10,\b+2) -- (\xend+10,\b+2);
  \node[name] at (-14,\b+1.5) {\textbf{(a) INT-$n$}\\ GPTQ, GPTAQ, AWQ};
  \node[info] at (\xend+14,\b+1.5) {one grid; same precision for every weight\\ stores $s,z$ per group-row};

  \def\b{-6.5}
  \foreach \g in {0,1,2,3} \draw[code] (\g*140,\b) rectangle ++(128,2);
  \foreach \p in {37, 230, 296, 381, 503} \fill[wsOut] (\p,\b) rectangle ++(3,3.5);
  \node[anchor=south] at (232,\b+3.5) {FP16};
  \draw[rate] (-10,\b+2) -- (\xend+10,\b+2);
  \node[name] at (-14,\b+1.5) {\textbf{(b) Outliers}\\ OWQ, SpQR, PB-LLM};
  \node[info] at (\xend+14,\b+1.5) {few weights kept in higher precision\\ stores their indices and a sparse path};

  \def\b{-13}
  \foreach \g/\h/\t/\ly in {0/3/{3 bits}/3,1/2/{2 bits}/2,2/1/{1 bit}/2,3/2/{2 bits}/2} {
    \draw[code] (\g*140,\b) rectangle ++(128,\h);
    \node[anchor=south] at (\g*140+64,\b+\ly) {\t};
  }
  \draw[rate] (-10,\b+2) -- (\xend+10,\b+2);
  \node[name] at (-14,\b+1.5) {\textbf{(c) Mixed precision}\\ SliM-LLM};
  \node[info] at (\xend+14,\b+1.5) {precision varies between groups only\\ stores a 2-bit width per group};

  \def\wsgroup#1#2#3#4{%
    \pgfmathsetmacro{\xm}{#1+#3}\pgfmathsetmacro{\xl}{#1+128-#4}
    \ifdim #3pt>0pt \draw[hi] (#1,#2) rectangle (\xm,#2+3); \fi
    \draw[mid] (\xm,#2) rectangle (\xl,#2+2);
    \ifdim #4pt>0pt
      \draw[lo] (\xl,#2) rectangle (#1+128,#2+1);
      \draw[shc] (\xl,#2+1) rectangle (#1+128,#2+2);
    \fi
    \draw[shf] (\xm,#2+2) rectangle (#1+128,#2+3);
  }

  \def\b{-19.5}
  \foreach \g in {0,1,2,3} \wsgroup{\g*140}{\b}{8}{16};
  \draw[rate] (-10,\b+2) -- (\xend+10,\b+2);
  \node[name] at (-14,\b+1.5) {\textbf{(d) \weshareone{}}\\ one layout, U8 D16};
  \node[info] at (\xend+14,\b+1.5) {precision varies within every group\\ stores $s,z$, flag $f$ and pair bits $c$;\\ bins at fixed positions: fixed wiring};

  \def\b{-26}
  \foreach \g/\u/\d/\o in {0/32/0/{+32},1/8/16/{0},2/0/64/{-32},3/4/8/{0}} {
    \wsgroup{\g*140}{\b}{\u}{\d}
    \node[anchor=north] at (\g*140+64,\b) {U\u{} D\d{}, $\delta=\o$};
  }
  \draw[rate] (-10,\b+2) -- (\xend+10,\b+2);
  \node[name] at (-14,\b+1.5) {\textbf{(e) \wesharetwo{}}\\ a layout per group};
  \node[info] at (\xend+14,\b+1.5) {bits move between groups, $\sum\delta\le 0$\\ adds a 6-bit layout index per group};

  \def\b{-31.6}
  \draw[code] (0,\b) rectangle ++(14,1.2);   \node[anchor=west] at (18,\b+0.6) {code bits stored per weight};
  \fill[wsOut] (232,\b) rectangle ++(4,1.2); \node[anchor=west] at (240,\b+0.6) {FP16 outlier};
  \draw[rate] (365,\b+0.6) -- ++(22,0);      \node[anchor=west] at (391,\b+0.6) {2 bits per weight};
  \def\b{-33.6}
  \draw[hi]  (0,\b) rectangle ++(14,1.2);    \node[anchor=west] at (18,\b+0.6) {high bin: $n$ bits};
  \draw[mid] (148,\b) rectangle ++(14,1.2);  \node[anchor=west] at (166,\b+0.6) {middle: $n{-}1$};
  \draw[lo]  (305,\b) rectangle ++(14,1.2);  \node[anchor=west] at (323,\b+0.6) {low: $n{-}2$};
  \draw[shf] (445,\b) rectangle ++(14,1.2);  \node[anchor=west] at (463,\b+0.6) {shared flag $f$ ($b_0$)};
  \draw[shc] (615,\b) rectangle ++(14,1.2);  \node[anchor=west] at (633,\b+0.6) {shared pair bit $c$ ($b_1$)};
  \node[anchor=north west,align=left,text width=158mm,inner sep=0pt] at (-205,-35.0) {%
    \textbf{Notation.} $n$: bits of a full code ($n=3$ here). $s,z$: scale and zero point of a group-row.
    $b_0,b_1$: two lowest bits of a code. $f$: flag, one stored bit per group-row, gives $b_0$ of every middle- and
    low-bin code. $c$: pair bit, one stored bit per two low-bin weights, gives their $b_1$. $U,D$: sizes of the high and
    low bins. $\delta=U-D/2$: offset of a layout in bits per group-row.};
\end{tikzpicture}
\caption{Overview of weight-only quantization for LLMs: How different formats store one row of a layer at 2 bits per
weight with four groups of 128 positions. Bar height is the code bits per weight. (a)-(c) store every code bit with its weight. In (d)-(e), \weshare{} uses a grid of $n=3$ bits:
the high, middle and low bins store $n$, $n-1$ and $n-2$ bits of each code, and the hatched bits are shared (Figure~\ref{fig:format}).
\weshareone{} uses U8 D16 in every group;
\wesharetwo{} picks a layout per group with $\sum\delta\le0$ per layer.}
\label{fig:sharing}
\end{figure*}

Serving large language models is limited by their weights. At small batch sizes, every weight is read from memory for
each generated token, and weight traffic far exceeds arithmetic~\cite{lin2024awq}, so fewer bits per weight mean faster
decoding and larger models per device. Post-training quantization usually stores weights in group-wise integer formats,
which split each row into groups of 128 input channels with a scale, a zero point and one integer code per weight. These
formats are close to lossless at 4 bits but degrade at 3 and especially at 2 bits. At 2 bits, GPTQ with a tuned clipping
range raises the WikiText-2 perplexity of Llama-3-8B and Qwen2.5-0.5B by factors of 6.4 and 9.9.

Most low-bit methods keep this format and improve the encoder~\cite{frantar2023gptq,li2025gptaq,shao2024omniquant,ma2024affinequant},
so every weight of a group maintains the same precision, although the input channels of a group affect the layer output very
unequally. Outlier methods~\cite{lee2024owq,dettmers2024spqr,yuan2024pbllm} keep a few weights in higher precision but
must store their positions. SliM-LLM~\cite{huang2025slimllm} varies precision between groups, which requires a width per
group and lowers decoding throughput, while still treating all weights of a group alike. Codebook and trellis
formats~\cite{tseng2024quipsharp,egiazarian2024aqlm,tseng2024qtip} achieve lower 2-bit perplexity but decode through
lookups or a trellis and often need fine-tuning. Varying precision within a group is harder, as a per-weight width
costs at least one bit per weight, half the budget of a 2-bit format.

Work on bit-level structure suggests another route. Accelerators prune the low-order bits of a group to a shared
constant~\cite{chen2024bbs}, RRAM designs store identical bit columns once~\cite{yang2025bitreorder}, and texture
formats, test compression and bus coding tie several values to one stored bit chosen by the encoder
(Section~\ref{sec:background}). The low bits of a weight carry little of its value, and even less in the less sensitive
channels of a group, so these weights can read their low bits from shared stored bits. Sharing constrains a weight
rather than discarding a bit. A weight that shares its lowest bit can still take any code of that parity, a grid with
twice the step whose offset is chosen per group-row. On the INT3 grid, this step is still 14\% finer than that of INT2.

We present \weshare{}, a format that keeps the integer grid and the dequantizer of group-wise quantization but varies
precision within every group (Figure~\ref{fig:sharing}). It works on a grid one bit finer than its rate: INT4 at 3 bits
and INT3 at 2 bits, and it places the input channels of each group into three bins at fixed positions determined by the Hessian
diagonal that GPTQ already computes. The high, middle and low bins store $n$, $n-1$ and $n-2$ bits of each code. One
stored flag per group-row supplies the lowest bit of every middle- and low-bin code, and one stored bit per pair of
low-bin weights supplies their second bit. This wiring is the same for every layer and model, and a sharing-aware GPTQ
chooses the shared values. The decoder needs no per-weight metadata or codebook. It routes bits into codes and runs
the unchanged INT-$n$ dequantizer. \weshareone{} gives every group the same layout, so the routing is fixed wiring.
\wesharetwo{} picks a layout per group within each layer's bit budget, incurring a 6-bit index
per group and a larger decoder. Both store 3.164 and 2.156 bits per weight, 0.016 more than INT3 and INT2.

\begin{figure*}[!ht]
\centering
\includegraphics[width=0.99\textwidth]{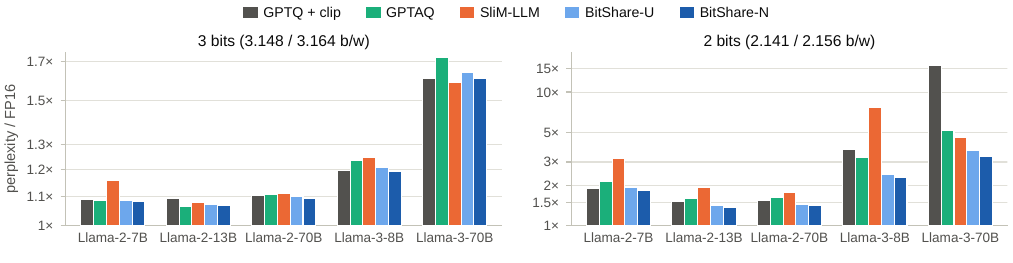}
\caption{WikiText-2 perplexity relative to FP16 (log scale) at 3 and 2 bits:  SliM-LLM and GPTAQ, GPTQ + clip and the two modes of \weshare{} (Llama-3-8B with 4096-token windows, the other models with 2048).}
\label{fig:methods}
\end{figure*}

\paragraph{Contributions} We introduce static bit sharing for LLM weights, in which the low bits of less sensitive
weights are read from stored bits through wiring fixed for every layer and model, and sensitivity-aware binning, which
varies precision within every group without per-weight metadata (Sections~\ref{sec:binning} and~\ref{sec:sharing}). To
our knowledge, this is the first integer weight format for LLMs whose sharing structure is fixed in advance and whose
shared values are chosen by the encoder. We define two modes, \weshareone{} with fixed wiring and \wesharetwo{} with
per-group layouts allocated exactly by dynamic programming (Section~\ref{sec:layouts}), and a sharing-aware GPTQ that
chooses every stored value, including the shared bits, with no fine-tuning or codebook (Section~\ref{sec:encoder}). We
evaluate both modes on eight LLMs against SliM-LLM, GPTAQ and GPTQ at equal storage, at equal perplexity and down to
collapse, with ablations of each part of the format (Section~\ref{sec:results}), and synthesize the decoder
(Section~\ref{sec:hw_impl}). At 2 bits, \wesharetwo{} lowers perplexity by 19-70\% against SliM-LLM and by
12.5-35.3\% against GPTAQ, and at SliM-LLM's perplexity \weshareone{} stores 13.6-19.4\% fewer bits per weight.
Figure~\ref{fig:methods} compares the perplexity of SliM-LLM, GPTAQ, GPTQ + clip and both modes of \weshare{} on WikiText-2 relative to FP16 at 3 and 2 bits, showing the efficacy of our proposed approach.

\providecommand{\cmark}{$\checkmark$}
\providecommand{\xmark}{$\times$}

\section{Background}
\label{sec:background}

\subsection{Weight-only quantization of LLMs}
\label{sec:bg-ptq}

\paragraph{Encoders on a fixed grid} Group-wise integer formats store a scale and a zero point per group of 128 input
channels and one code per weight (Section~\ref{sec:prelim}), so all weights of a group share one precision.
GPTQ~\cite{frantar2023gptq}, derived from OBS~\cite{hassibi1993obs} and OBQ~\cite{frantar2022obc}, rounds one column
at a time and compensates the error in the remaining columns through the inverse Hessian. GPTAQ~\cite{li2025gptaq}
also matches each layer's output to the full-precision model, OmniQuant~\cite{shao2024omniquant} and
AffineQuant~\cite{ma2024affinequant} learn quantizer parameters by gradient descent and QuaRot~\cite{ashkboos2024quarot}
removes outliers with rotations. All of them keep the INT-$n$ storage format.

\paragraph{Codebook and trellis formats} QuIP\#~\cite{tseng2024quipsharp}, AQLM~\cite{egiazarian2024aqlm} and
QTIP~\cite{tseng2024qtip} reach lower 2-bit perplexity with lattice or additive codebooks or a trellis code, usually
with Hadamard transforms and fine-tuning, and decode by lookup or from trellis states. In QTIP, neighboring weights
share stored bits through overlapping trellis windows.

\subsection{Salience and mixed precision}
\label{sec:bg-salience}

The error of weight $w_{ij}$ reaches the output scaled by the activation energy of input channel $j$, the diagonal of
$\mathbf{H}=2\mathbf{X}\mathbf{X}^{\top}$. %
AWQ~\cite{lin2024awq} found that keeping 1\% of the channels in FP16, selected by activation magnitude, lowers the round-to-nearest INT3 perplexity of OPT-6.7B from 23.54 to 11.39, but rescales
salient channels instead to keep one precision.
OWQ~\cite{lee2024owq}, SpQR~\cite{dettmers2024spqr}, SqueezeLLM~\cite{kim2024squeezellm} and PB-LLM~\cite{yuan2024pbllm}
keep outliers in higher precision at the cost of stored positions and a sparse decode path. HAWQ~\cite{dong2019hawq}
assigns bit widths per layer. SliM-LLM~\cite{huang2025slimllm} assigns $N-1$, $N$ or $N+1$ bits per group,
stores a 2-bit width per group and reports 20-27\% lower decode throughput than GPTQ at 2 bits
(Section~\ref{sec:cost}). Precision is uniform within each group.

\subsection{Bit-level techniques and shared bits}
\label{sec:bg-bits}

Bit-serial accelerators such as Stripes~\cite{judd2016stripes}, Pragmatic~\cite{albericio2017pragmatic},
Bitlet~\cite{lu2021bitlet} and BitL~\cite{lee2025bitl} skip or reorder zero bits to speed up arithmetic on stored
values. BBS~\cite{chen2024bbs} also prunes the low-order bits of a weight group to a shared constant.
Yang et al.~\cite{yang2025bitreorder} reorder RRAM crossbar rows so that pairs of bit columns become identical and store one column
per pair, which is lossless but depends on the weight values and needs stored indices. Shannonic~\cite{ibrahim2026shannonic}
entropy-codes quantized tensors but cannot go below the entropy of the codes, and microscaling
formats~\cite{rouhani2023mx} share an exponent per block, as the scale already does in group-wise formats. Outside
machine learning, a fixed sharing structure with encoder-chosen values is common. BC7 texture
compression~\cite{microsoft2024bc7} stores a P-bit shared as the least significant bit of several color components,
Illinois Scan~\cite{banerjee2010illinois} and EDT~\cite{rajski2004edt} expand a few tester bits into many scan cells,
bus-invert coding~\cite{stan1995businvert} sets one invert bit per group of bus lines and HIPR~\cite{dasgupta2025hipr}
shares configuration bits between neighboring LUTs. A bit shared by two values is the simplest case of trellis-coded
quantization~\cite{marcellin1990tcq}.

\begin{table}[!htbp]
\centering
\scriptsize
\setlength{\tabcolsep}{2.6pt}
\providecommand{\rh}[1]{\rotatebox{90}{#1}}
\caption{Comparing weight-only quantization formats for LLMs.}
\label{tab:related}
\begin{tabular}{@{}lccccccc@{}}
\toprule
Method & \rh{Codebook-free?} & \rh{Index-free?} & \rh{Single dequantizer?} & \rh{Sensitivity-aware binning?} &
  \rh{Intra-group precision?} & \rh{Per-group layout?} & \rh{Shared bits?} \\
\midrule
GPTQ, GPTAQ, AWQ           & \cmark & \cmark & \cmark & \xmark & \xmark & \xmark & \xmark \\
OmniQuant, AffineQuant     & \cmark & \cmark & \cmark & \xmark & \xmark & \xmark & \xmark \\
OWQ, SpQR, PB-LLM          & \cmark & \xmark & \xmark & \cmark & \cmark & \xmark & \xmark \\
SliM-LLM                   & \cmark & \cmark & \xmark & \cmark & \xmark & \cmark & \xmark \\
QuIP\#, AQLM               & \xmark & \cmark & \cmark & \xmark & \xmark & \xmark & \xmark \\
QTIP                       & \xmark & \cmark & \cmark & \xmark & \xmark & \xmark & \cmark \\
\midrule
\weshareone{} (ours)       & \cmark & \cmark & \cmark & \cmark & \cmark & \xmark & \cmark \\
\wesharetwo{} (ours)       & \cmark & \cmark & \cmark & \cmark & \cmark & \cmark & \cmark \\
\bottomrule
\end{tabular}
\end{table}

\subsection{Motivation for \weshare{}}
\label{sec:bg-motivation}

Table~\ref{tab:related} summarizes prior formats. Integer formats decode cheaply but give
every weight in a group the same precision. Formats that vary precision pay with stored positions and a sparse path, or
with a per-group width map that still leaves each group on one grid. Codebook and trellis formats give up the integer
dequantizer, and bit-level techniques share only bits that happen to be equal after quantization. To our knowledge, no
integer weight format for LLMs fixes a sharing structure in advance and lets a second-order encoder choose the shared
values. \weshare{} fills this gap. It varies precision within every group at fixed positions, without per-weight
metadata, and pays for it with shared bits chosen by GPTQ, while keeping the INT-$n$ dequantizer. Our scope is
post-training scalar integer formats without gradient-based training, the setting of SliM-LLM's comparison
(Section~\ref{sec:setup}).

\providecolor{wsLine}{RGB}{0,0,0}
\providecolor{wsHatch}{RGB}{31,78,121}
\providecolor{wsHigh}{RGB}{33,113,181}
\providecolor{wsMid}{RGB}{239,243,255}
\providecolor{wsLow}{RGB}{107,174,214}
\usetikzlibrary{patterns}

\section{Methodology}
\label{sec:method}

\weshare{} stores every linear layer on the integer grid of standard group-wise quantization, but gives the weights of a
group different numbers of stored bits. Sensitivity-aware binning places the input channels of each
group into three bins. A weight in a less sensitive bin stores fewer bits of its code, and every bit it does not store is
\emph{shared}: it is tied by fixed wiring to one stored bit that several weights read. {The sizes
of a group's bins form its \emph{layout}. \weshare{} has two modes, which differ only in how the groups get their layouts:
\weshareone{} (uniform) gives every group of every layer the same layout, and \wesharetwo{} (non-uniform) gives each group a
layout from a fixed menu, so that groups that need more bits take them from groups that need fewer at the same total. In
both modes a sharing-aware version of GPTQ chooses every stored value.}

\subsection{Group-wise integer quantization}
\label{sec:prelim}

A linear layer computes $\mathbf{y}=\mathbf{W}\mathbf{x}$ with $\mathbf{W}\in\mathbb{R}^{R\times C}$. Group-wise
quantization splits each row into groups of $G=128$ consecutive input channels and gives every \emph{group-row} $(i,g)$ a
scale $s_{ig}$ and a zero point $z_{ig}\in\{0,\dots,2^n-1\}$ on an $n$-bit grid. Weight $w_{ij}$ is stored as an $n$-bit
code $q_{ij}$ and reconstructed as
\begin{equation}
  \hat w_{ij} \;=\; s_{ig}\,\bigl(q_{ij}-z_{ig}\bigr).
  \label{eq:dequant}
\end{equation}
With the asymmetric min-max convention and a clipping factor $p\le 1$,
$s_{ig}=p\,(\max_{j} w_{ij}-\min_{j} w_{ij})/(2^n-1)$ and $z_{ig}=\mathrm{round}(-p\min_{j} w_{ij}/s_{ig})$, clamped to
the grid, with the extremes taken over the group. INT-$n$ stores $nG$ code bits, a 16-bit scale and an $n$-bit zero point
per group-row, $n+(16+n)/G$ bits per weight: 3.148 for INT3 and 2.141 for INT2 at $G=128$.

\weshare{} keeps the reconstruction of Equation~\ref{eq:dequant} as it is and works on a grid one bit finer than its rate: the 3-bit format uses the
INT4 grid ($n=4$) and the 2-bit format the INT3 grid ($n=3$). Writing a code by its bits, $q=\sum_{k=0}^{n-1}b_k2^k$,
each weight stores the high bits of its code and takes some of its low bits from shared bits.

\subsection{Sensitivity-aware binning}
\label{sec:binning}

The error of weight $w_{ij}$ reaches the layer output, scaled by the activation of input channel $j$, so the channels of a
group matter unequally. We measure the sensitivity of channel $j$ by the diagonal of the Hessian of the layer's output
error, $h_j=[\mathbf{H}]_{jj}$ with $\mathbf{H}=\frac{2}{N}\mathbf{X}\mathbf{X}^{\top}$ over $N$ calibration tokens, the
matrix GPTQ already computes~\cite{frantar2023gptq}.

Group $g$ has a layout $(U_g,D_g)$: its $U_g$ most sensitive channels form the \emph{high bin} $\mathcal{H}_g$, its $D_g$
least sensitive form the \emph{low bin} $\mathcal{L}_g$ and the remaining $M_g=G-U_g-D_g$ form the \emph{middle bin}
$\mathcal{M}_g$. The bins occupy fixed positions within the group: positions $0,\dots,U_g-1$ hold the high bin,
$U_g,\dots,G-D_g-1$ the middle bin and $G-D_g,\dots,G-1$ the low bin, each bin keeping the original order of its
channels. A position's bin, and with it which of its bits are stored and which are shared, thus depends only on the
position and on the two numbers $U_g$ and $D_g$. Binning is a permutation $\pi$ of the input channels within each group,
one per layer and shared by all $R$ rows.

The permutation is applied to the columns of $\mathbf{W}$ offline. At inference the activations are read through it, or
equivalently the decoded rows are written through its inverse. It costs one 7-bit index per input channel and layer,
shared by all rows. In \wesharetwo{} every linear layer has its own, since its layouts are its own, while
in \weshareone{} the layers that read the same input share one. Counting one per linear layer, the permutations add
0.0012 bits per weight for Llama-2-7B.

\subsection{Static bit sharing}
\label{sec:sharing}

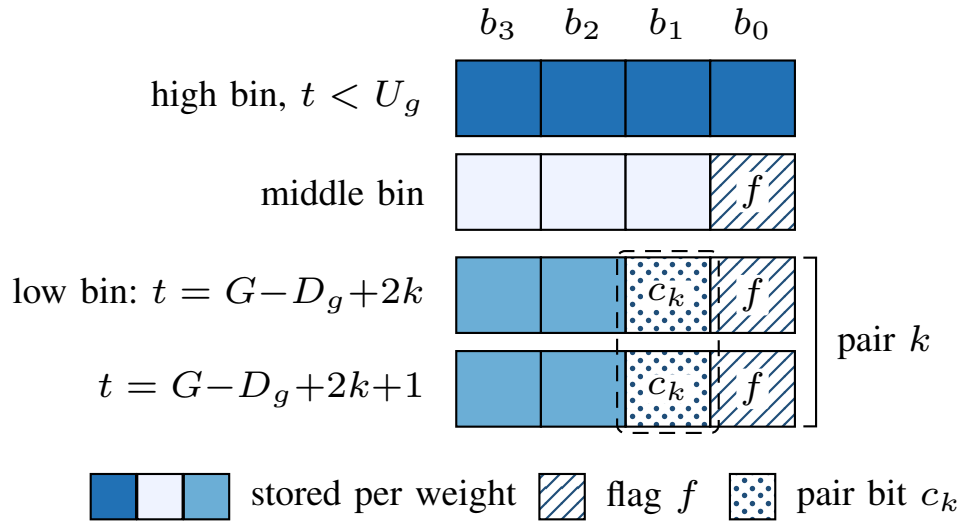
\begin{figure}[t]
\centering
\begin{tikzpicture}[x=5.6mm,y=6.2mm,
  bit/.style={draw=wsLine,minimum width=5.6mm,minimum height=5mm,inner sep=0pt,font=\scriptsize},
  hi/.style={bit,fill=wsHigh},
  mid/.style={bit,fill=wsMid},
  lo/.style={bit,fill=wsLow},
  shf/.style={bit,pattern=north east lines,pattern color=wsHatch},
  shc/.style={bit,pattern=crosshatch dots,pattern color=wsHatch},
  tag/.style={fill=white,inner sep=0.7pt,rounded corners=0.6pt,font=\scriptsize},
  key/.style={minimum width=3mm,minimum height=3mm},
  lab/.style={anchor=east,font=\scriptsize,align=right}]
  \foreach \k/\x in {3/0,2/1,1/2,0/3} \node[font=\scriptsize] at (\x,4.8) {$b_{\k}$};
  \node[lab] at (-0.65,4) {high bin, $t<U_g$};
  \foreach \x in {0,1,2,3} \node[hi] at (\x,4) {};
  \node[lab] at (-0.65,3) {middle bin};
  \foreach \x in {0,1,2} \node[mid] at (\x,3) {};
  \node[shf] at (3,3) {}; \node[tag] at (3,3) {$f$};
  \node[lab] at (-0.65,1.9) {low bin: $t=G{-}D_g{+}2k$};
  \node[lab] at (-0.65,0.9) {$t=G{-}D_g{+}2k{+}1$};
  \foreach \y in {1.9,0.9} {
    \node[lo] at (0,\y) {}; \node[lo] at (1,\y) {};
    \node[shc] at (2,\y) {}; \node[tag] at (2,\y) {$c_k$};
    \node[shf] at (3,\y) {}; \node[tag] at (3,\y) {$f$};
  }
  \draw[densely dashed,rounded corners=1.5pt] (1.4,2.37) rectangle (2.6,0.43);
  \draw (3.6,2.3) -- (3.75,2.3) -- (3.75,0.5) -- (3.6,0.5);
  \node[anchor=west,font=\scriptsize] at (3.78,1.4) {pair $k$};
  \node[hi,key] at (-4.55,-0.25) {}; \node[mid,key] at (-4.0,-0.25) {}; \node[lo,key] at (-3.45,-0.25) {};
  \node[anchor=west,font=\scriptsize] at (-3.15,-0.25) {stored per weight};
  \node[shf,key] at (0.75,-0.25) {};
  \node[anchor=west,font=\scriptsize] at (1.05,-0.25) {flag $f$};
  \node[shc,key] at (3.0,-0.25) {};
  \node[anchor=west,font=\scriptsize] at (3.3,-0.25) {pair bit $c_k$};
\end{tikzpicture}
\caption{The 3-bit format: the bits of the INT4 code $q=b_3b_2b_1b_0$ of a weight in each bin of group $g$. Solid
bits are stored with the weight, in the color of its bin. Patterned bits are read from shared stored bits: every $f$ is
the one flag $f_{ig}$ of the group-row, and the two $c_k$ of a low-bin pair are its one pair bit $c_{igk}$. A group-row also stores its scale $s_{ig}$
and zero point $z_{ig}$. The 2-bit format drops $b_3$ in every bin (INT3 grid).}
\label{fig:format}
\end{figure}

The high, middle and low bins store $n$, $n-1$ and $n-2$ bits of each code, always the most significant ones. The
missing low bits come from two kinds of shared bits (Figure~\ref{fig:format}):
\begin{itemize}
  \item the \emph{pair bit} $c_{igk}$: the low-bin positions are paired as $(G-D_g+2k,\,G-D_g+2k+1)$,
        $0\le k<D_g/2$, and both weights of a pair read their bit $b_1$ from one stored bit;
  \item the \emph{flag} $f_{ig}$: bit $b_0$ of every middle- and low-bin weight of group-row $(i,g)$ is one stored bit.
\end{itemize}
The code of the weight at position $t$ of group-row $(i,g)$ is thus
\begin{equation}
  q \;=\;
  \begin{cases}
    v                                & t<U_g, \\
    2v + f_{ig}                      & U_g\le t<G-D_g, \\
    4v + 2c_{igk(t)} + f_{ig}        & t\ge G-D_g,
  \end{cases}
  \label{eq:codes}
\end{equation}
where $v$ is the weight's stored field of $n$, $n-1$ or $n-2$ bits and $k(t)=\lfloor (t-G+D_g)/2\rfloor$ is its pair.

Sharing constrains the codes rather than discarding bits. A middle-bin weight can take any code of parity $f_{ig}$: an
$(n-1)$-bit grid with twice the fine step. A low-bin weight can take any code congruent to $f_{ig}+2c_{igk}$ modulo 4: a
grid with four times the fine step whose offset is chosen per pair. All shared bits are programmable. The encoder sets
each pair bit, for every row, to the value that minimizes the pair's error as weighted in the layer output, so that when
the two weights need different values the bit goes to the one whose error counts more. It also sets the flag of
each group-row. A bit dropped from the code, in contrast, is the same for every weight that lacks it.

The flag lets the encoder choose, for one bit per group-row, which end of the group's range the narrower
grids of the middle and low bins leave uncovered, rather than tying it to the parity of the zero point
(Appendix~\ref{app:format}).

A group-row with layout $(U,D)$ holds $nU+(n-1)M+(n-2)D+D/2$ code and pair bits, that is
\begin{equation}
  (n-1)G+\delta(U,D),\qquad \delta(U,D)=U-\tfrac{D}{2},
  \label{eq:rate}
\end{equation}
so the extra bits of the high bin are paid by the low bin at two positions per bit. The \emph{reference layout}
$U=8$, $D=16$ ($M=104$) has $\delta=0$: exactly the payload of INT-$(n-1)$. \weshareone{} uses it in every group.
Layouts with $\delta<0$ store fewer bits. Sharing $b_1$ among $B>2$ low-bin positions, or dropping it, lowers the
rate further (Appendix~\ref{app:format}). Both modes use pairs ($B=2$), and Sections~\ref{sec:storage}
and~\ref{sec:collapse} use the other variants only to extend the rate-quality curve of \weshareone{}.

\subsection{A layout per group \texorpdfstring{(\wesharetwo{})}{(BitShare-N)}}
\label{sec:layouts}

The one layout of \weshareone{} fits the average group, but groups differ in how their sensitivity is spread over
their channels, in its overall level and in their range, so the same bits buy more accuracy in some groups than in
others. \wesharetwo{} therefore lets each group take its layout from a menu $\Lambda$ of 44 layouts, the same for
every layer and model. The menu contains every combination of $U\in\{0,4,8,12,16,24,32\}$ and
$D\in\{0,16,32,48,64,96\}$ as well as the layouts with
$\delta(U,D)=0$ ($U=D/2$, from U0~D0 to U32~D64), all with pairs in the low bin. Their offsets $\delta$ range from $-48$ to
$+32$ bits per group-row (Figure~\ref{fig:sharing}). \weshareone{} is the special case of a menu with one layout,
$\Lambda=\{$U8~D16$\}$ unless a lower rate is wanted.

For every group $g$ and layout $\ell\in\Lambda$, the encoder measures the selection error
\begin{equation}
  E_g(\ell) \;=\; \sum_{i=1}^{R}\sum_{j\in g} h_j\,\bigl(w_{ij}-\hat w^{(\ell)}_{ij}\bigr)^2,
  \label{eq:selerr}
\end{equation}
where $\hat w^{(\ell)}$ rounds each weight to the nearest code that its bin allows under layout $\ell$, with the scale, zero
point and flag of each group-row found by the same range search that sets the group parameters (Section~\ref{sec:encoder}), and
each pair bit taking the value with the smaller error. The layouts of a layer then solve
\begin{equation}
  \min_{\ell_1,\dots,\ell_{C/G}\in\Lambda}\ \sum_g E_g(\ell_g)
  \quad\text{subject to}\quad \sum_g \delta(\ell_g)\le 0,
  \label{eq:alloc}
\end{equation}
so the layer stores at most the reference payload, $(n-1)$ bits per weight, and a group can take bits from another only
where they lower the layer's error. Equation~\ref{eq:alloc} is a multiple-choice knapsack problem over the layer's groups.
Since all offsets are multiples of 4, dynamic programming over the running sum of the offsets solves it exactly with at most
$20\,C/G+1$ states. Each layer solves Equation~\ref{eq:alloc} with its own sensitivities, so the layouts differ between layers as well
as between the groups of a layer. The layout is
stored as one 6-bit index per group, shared by all $R$ rows of the layer, which is $6/(GR)\approx10^{-5}$ bits per weight
for $R\ge4096$. \weshareone{} stores no index.

\subsection{Storage format and decoder}
\label{sec:decoder}

\begin{table}[t]
\caption{Bits per group-row ($G=128$, pairs) for a group with layout $(U,D)$ and for the reference layout U8~D16
($M=104$), which \weshareone{} uses in every group, and bits per weight. In \wesharetwo{}, a layer's offsets $\delta$
sum to at most 0, so its total is at most the reference's. The last row gives the bits per weight of INT3 and INT2 with
the same group size, scale and zero point.}
\label{tab:budget}
\centering
\small
\setlength{\tabcolsep}{4pt}
\begin{tabular}{@{}lccc@{}}
\toprule
Field                                   & layout $(U,D)$  & \multicolumn{2}{c}{U8~D16} \\
                                        &                 & $n=4$ & $n=3$ \\
\midrule
High-bin codes                          & $nU$            & 32    & 24    \\
Middle-bin fields                       & $(n-1)M$        & 312   & 208   \\
Low-bin fields                          & $(n-2)D$        & 32    & 16    \\
Pair bits                               & $D/2$           & 8     & 8     \\
Scale (FP16)                            & 16              & 16    & 16    \\
Zero point                              & $n$             & 4     & 3     \\
Flag                                    & 1               & 1     & 1     \\
\midrule
Total bits                              & $(n{-}1)G{+}\delta{+}n{+}17$ & 405 & 276 \\
Bits per weight                         &                 & 3.164 & 2.156 \\
\midrule
INT3 / INT2                             &                 & 3.148 & 2.141 \\
\bottomrule
\end{tabular}
\end{table}

Table~\ref{tab:budget} lists the fields of a group-row. In \weshareone{} with U8~D16, and in \wesharetwo{} when the
offsets of a layer sum to zero, the code and pair bits of a layer equal those of INT-$(n-1)$. On top of these, the
format stores the flag and one more zero-point bit, since $z_{ig}$ is a code of the finer grid. This is $2/128=0.016$ bits per weight over
INT-$(n-1)$, for totals of 3.164 and 2.156 bits per weight in both modes. The layout indices of \wesharetwo{} add about
$10^{-5}$.

A group-row decodes from its own fields and its group's layout $(U_g,D_g)$, which in \weshareone{} is the same
for every group and in \wesharetwo{} is looked up from the group's 6-bit index in the menu, a table of 44 entries shared
by every layer and model. The decoder reads $s_{ig}$, $z_{ig}$ and $f_{ig}$, then the $U_g$ $n$-bit, $M_g$ $(n-1)$-bit and
$D_g$ $(n-2)$-bit fields and the $D_g/2$ pair bits, forms the codes by Equation~\ref{eq:codes} and dequantizes them
with Equation~\ref{eq:dequant}, exactly as INT-$n$ does. Forming the codes needs no arithmetic: a position's bin follows from two
comparisons with its group's boundaries, which in \weshareone{} are the same in every group, so that the routing is fixed
wiring. In \wesharetwo{}, group-rows of different layouts have different sizes, but every row of a layer has the same
sequence of layouts, so a group's offset within a row is a prefix sum over the layer's layout indices, computed once per
layer. The format has no
per-weight or per-block metadata and no codebook or look-up table of weight values.

\subsection{Sharing-aware GPTQ}
\label{sec:encoder}

The encoder chooses every stored value. Like GPTQ~\cite{frantar2023gptq}, it quantizes the transformer blocks in order on
calibration data and, within a block, the linear layers in execution order, each on the inputs produced by the layers
already quantized. For a layer with inputs $\mathbf{X}$, GPTQ minimizes $\|(\mathbf{W}-\hat{\mathbf{W}})\mathbf{X}\|_F^2$
column by column: after a column is quantized, its error is compensated in the columns not yet quantized through the
inverse Hessian. \wesharetwo{} first chooses the layouts of the layer, while \weshareone{} gives
every group its one layout. Both modes then change three steps of GPTQ, as listed in Algorithm~\ref{alg:encoder}
(Appendix~\ref{app:algo}).

\paragraph{Group parameters with the format in the loop} Before the column loop, the scale, zero point and flag of every
group-row are fixed, as in GPTQ with static groups. For each clipping factor $p\in\{1.00,0.98,\dots,0.80\}$ and each flag
value $f\in\{0,1\}$, every weight of the group-row is rounded to the nearest code that its bin allows, each low-bin pair
taking the pair bit with the smaller error. The candidate with the smallest sensitivity-weighted error
$\sum_{j}h_j(w_{ij}-\hat w_{ij})^2$ is kept. The flag is thus chosen jointly with the range, which doubles the 11-point
range search. The selection error (Equation~\ref{eq:selerr}) of a layout is this search's error, summed over the rows.

\paragraph{Units in order of sensitivity} GPTQ quantizes \emph{units}: a single column in the high and middle bins
and a pair of columns in the low bin (a block of $B$ columns for a bit shared by $B$, single columns for a dropped bit).
The units are quantized in decreasing order of sensitivity over the whole layer (act-order), and a pair or block ranks by
its most sensitive column.
The most sensitive channels are therefore quantized first, while the most columns remain to
compensate their error.

\paragraph{Shared bits chosen by the GPTQ objective} A single column is rounded, row by row, to the nearest code its bin
allows given the flag. For a pair, both values $c$ of the pair bit are tried in every row. The first column is
rounded to the nearest code with $b_1=c$, and its error is propagated to the second column, as GPTQ propagates it to any
later column, before the second column is rounded. The value with the smaller objective $\sum_{j\in\text{unit}}(w_{ij}-\hat w_{ij})^2/d_j^2$
is kept, where $d_j$ is the diagonal of the Cholesky factor of $\mathbf{H}^{-1}$ that GPTQ uses to weigh errors. The
unit's errors are then propagated to the remaining columns.

\hyphenation{GPTAQ}

\begin{table*}[!ht]
\caption{WikiText-2 perplexity under SliM-LLM's protocol (calibration on WikiText-2), at 3 and 2 bits. Llama-3-8B is evaluated with 4096-token windows, as in SliM-LLM's paper, and the other models with 2048-token windows. Lowest perplexity values are in \textbf{bold}.}
\label{tab:slim}
\centering
\footnotesize
\setlength{\tabcolsep}{3pt}
\resizebox{0.95\textwidth}{!}{%
\begin{tabular*}{\textwidth}{@{\extracolsep{\fill}}lc rrr rr c c rrr rr@{}}
\toprule
 & \multicolumn{6}{c}{3 bits} & & \multicolumn{6}{c}{2 bits} \\
\cmidrule(lr){2-7}\cmidrule(l){9-14}
 & & \multicolumn{3}{c}{Llama-2} & \multicolumn{2}{c}{Llama-3} & & & \multicolumn{3}{c}{Llama-2} & \multicolumn{2}{c}{Llama-3} \\
\cmidrule(lr){3-5}\cmidrule(lr){6-7}\cmidrule(lr){10-12}\cmidrule(l){13-14}
Method & b/w & 7B & 13B & 70B & 8B & 70B & & b/w & 7B & 13B & 70B & 8B & 70B \\
\midrule
FP16 & 16 & 5.47 & 4.88 & 3.32 & 5.75 & 2.85 & & 16 & 5.47 & 4.88 & 3.32 & 5.75 & 2.85 \\
\midrule
RTN & 3.148 & 6.66 & 5.51 & 3.97 & 27.91$^\dagger$ & 11.84 & & 2.141 & 4.2e3 & 122.1 & 27.27 & 1.9e3$^\dagger$ & 4.6e5 \\
AWQ & 3.148 & 6.24 & 5.32 & -- & 8.22$^\dagger$ & 4.81 & & 2.141 & 2.2e5 & 1.2e5 & -- & 1.7e6$^\dagger$ & 1.7e6 \\
QuIP & -- & -- & -- & -- & -- & -- & & -- & 39.73 & 13.48 & 6.64 & 84.97$^\dagger$ & 13.03 \\
PB-LLM & -- & -- & -- & -- & -- & -- & & -- & 25.37 & 49.81 & NaN & 44.12$^\dagger$ & 11.68 \\
\addlinespace[2pt]
OmniQuant & 3.148 & 6.03 & 5.28 & 3.78 & -- & -- & & 2.141 & 11.06 & 8.26 & 6.55 & -- & -- \\
AffineQuant & 3.148 & 6.08 & 5.28 & -- & -- & -- & & 2.141 & 10.87 & 7.64 & -- & -- & -- \\
SliM-LLM+ & 3.148 & 5.94 & \textbf{5.11} & \textbf{3.35} & -- & -- & & 2.141 & 10.87 & 7.59 & 6.44 & -- & -- \\
\addlinespace[2pt]
SliM-LLM~{\bstars} & 3.148 & 6.33 & 5.26 & 3.69 & 7.16 & \textbf{4.53} & & 2.141 & 17.31 & 9.47 & 5.85 & 44.35 & 12.98 \\
GPTAQ~{\bstars} & 3.148 & 5.93 & 5.19 & 3.67 & 7.11 & 4.91 & & 2.141 & 11.76 & 7.84 & 5.39 & 18.76 & 14.62 \\
GPTQ + clip~{\bstars} & 3.148 & 5.96 & 5.33 & 3.66 & 6.88 & 4.58 & & 2.141 & 10.37 & 7.41 & 5.15 & 21.22 & 45.13 \\
\weshareone{} & 3.164 & 5.94 & 5.22 & 3.65 & 6.94 & 4.67 & & 2.156 & 10.53 & 6.88 & 4.76 & 13.87 & 10.34 \\
\wesharetwo{} & 3.164 & \textbf{5.92} & 5.21 & 3.63 & \textbf{6.85} & 4.58 & & 2.156 & \textbf{10.11} & \textbf{6.70} & \textbf{4.71} & \textbf{13.29} & \textbf{9.45} \\
\bottomrule
\end{tabular*}
}
\footnotesize\\
$^\dagger$~The Llama-3-8B baselines that SliM-LLM reports match 2048-token windows. b/w: bits per weight. --: not reported. \\
{\bstars}~{Our runs using released code for SliM-LLM~\cite{slimllm_github}, GPTAQ~\cite{gptaq_github} and GPTQ~\cite{gptq_github}. }
\end{table*}

\section{Results}
\label{sec:results}

\subsection{Evaluation Setup}
\label{sec:setup}

\noindent\textbf{Models.} Llama-2 (7B, 13B and 70B)~\cite{touvron2023llama2}, Llama-3 (8B and 70B)~\cite{grattafiori2024llama3}
and Qwen2.5 (0.5B, 1.5B and 3B)~\cite{yang2024qwen25}. All linear layers of the transformer blocks are quantized.
The embeddings and the output layer stay in FP16.

\noindent\textbf{Methods.} Every method quantizes groups of $G=128$ input channels with an FP16 scale and an integer zero point
per group-row.
\begin{itemize}
 
    \item \emph{GPTQ}~\cite{frantar2023gptq}: INT3 or INT2 with the min-max
        range, act-order with static groups, damping $0.01\,\overline{h}$ and
        sequential layers (Algorithm~\ref{alg:encoder}). On test layers our
        implementation produces the same codes as the released
        GPTQ \cite{gptq_github}.

    \item \emph{GPTQ + clip}: GPTQ with the range search of \weshare{},
        $p\in\{1.00,0.98,\dots,0.80\}$; at 2 bits also with $G=64$, which
        stores 0.125 bits per weight more than \weshare{}.

    \item \emph{SliM-LLM}~\cite{huang2025slimllm}: 2, 3 or 4 bits per group by
        salience, 3 on average per layer (1-3 bits, 2 on average, at 2 bits),
        with its own GPTQ. We run its released
        code \cite{slimllm_github}
        with the released group widths on Llama-2-7B/13B and Llama-3-8B and
        its own width search on the 70B models, for which none are released
        (Appendix~\ref{app:baselines}). We report our runs.

    \item \emph{GPTAQ}~\cite{li2025gptaq} adds to GPTQ's update a term that
        matches each layer's output to the full-precision model's,
        compensating the error of earlier layers. We run its released code \cite{gptaq_github}
        on the INT3 or INT2 grid with act-order and its own range search
        (Appendix~\ref{app:baselines}).

    \item \emph{Reported only}: under SliM-LLM's protocol we also list the
        perplexities SliM-LLM reports for RTN, AWQ~\cite{lin2024awq},
        QuIP~\cite{chee2023quip} and PB-LLM~\cite{yuan2024pbllm}, which need
        no gradient training, and for OmniQuant~\cite{shao2024omniquant},
        AffineQuant~\cite{ma2024affinequant} and SliM-LLM+, which learn their
        quantizer parameters by gradient descent.
        
  \item \weshare{}: the 3-bit format on the INT4 grid and the 2-bit format on the INT3 grid, in its two modes:
        \weshareone{}, with U8~D16 in every group unless stated otherwise, and \wesharetwo{}, with
        a layout per group chosen by Equation~\ref{eq:alloc}.
\end{itemize}

Bits per weight count every field of a group-row: 3.148 for INT3, 3-bit SliM-LLM and GPTAQ, and 2.141 for INT2, 2-bit SliM-LLM and GPTAQ. Both modes of \weshare{} store 3.164 and 2.156
(Table~\ref{tab:budget}).
The permutation
and, in \wesharetwo{}, the layout indices add 0.0006-0.0013 bits per weight on the Llama models and 0.002-0.005 on the
Qwen models. The group index that GPTAQ stores for each column when it forms groups in act-order adds about 0.001.
None of these bits is counted.

\noindent\textbf{Calibration and evaluation.} The comparison with SliM-LLM and GPTAQ and every run on the 70B models follow
SliM-LLM's protocol: 128 windows of 2048 tokens drawn from the WikiText-2 training split with GPTQ's loader and
SliM-LLM's seed. Across model families we also calibrate on 128 sequences of 2048 tokens from the validation split of
the Pile~\cite{gao2020pile}, built as in AWQ~\cite{lin2024awq}. We report perplexity on the WikiText-2 test split~\cite{merity2017pointer} in windows
of 2048 tokens and on the first 256 windows of 2048 tokens of GPTQ's C4 validation text~\cite{raffel2020t5}.
Appendix~\ref{app:results} adds WikiText-2 in windows of 4096 tokens. All runs use NVIDIA RTX A6000 GPUs.

\begin{table*}[!ht]
\centering
\caption{Perplexity ($\downarrow$) at equal storage, calibration on the Pile. b/w: bits per weight. Lowest perplexity values are in \textbf{bold}.}
\label{tab:main}
\footnotesize
\setlength{\tabcolsep}{1pt}
\resizebox{0.95\textwidth}{!}{%
\begin{tabular*}{\textwidth}{@{\extracolsep{\fill}}lc rrrrrr c rrrrrr@{}}
\toprule
 & & \multicolumn{6}{c}{WikiText-2} & & \multicolumn{6}{c}{C4} \\
\cmidrule(lr){3-8}\cmidrule(l){10-15}
 & & \multicolumn{2}{c}{Llama-2} & Llama-3 & \multicolumn{3}{c}{Qwen2.5} & & \multicolumn{2}{c}{Llama-2} & Llama-3 & \multicolumn{3}{c}{Qwen2.5} \\
\cmidrule(lr){3-4}\cmidrule(lr){5-5}\cmidrule(lr){6-8}\cmidrule(lr){10-11}\cmidrule(lr){12-12}\cmidrule(l){13-15}
Method & b/w & 7B & 13B & 8B & 0.5B & 1.5B & 3B & & 7B & 13B & 8B & 0.5B & 1.5B & 3B \\
\midrule
FP16 & 16 & 5.47 & 4.88 & 6.14 & 13.07 & 9.27 & 8.03 & & 7.26 & 6.73 & 9.44 & 20.38 & 15.12 & 13.36 \\
\midrule
\multicolumn{15}{@{}l}{\textit{3 bits}} \\
GPTQ & 3.148 & 6.26 & 5.41 & 7.96 & 19.70 & 12.13 & 10.01 & & 8.38 & 7.41 & 12.82 & 29.21 & 18.67 & 15.78 \\
GPTQ + clip & 3.148 & 6.06 & 5.45 & 7.82 & 19.11 & 11.23 & 9.45 & & 8.07 & 7.52 & 12.09 & 28.51 & 17.71 & 15.15 \\
\weshareone{} & 3.164 & \textbf{6.01} & 5.34 & 7.43 & 17.60 & \textbf{10.91} & 9.29 & & \textbf{8.02} & 7.30 & 11.72 & 26.43 & 17.39 & 14.94 \\
\wesharetwo{} & 3.164 & 6.02 & \textbf{5.29} & \textbf{7.41} & \textbf{17.25} & 10.94 & \textbf{9.26} & & \textbf{8.02} & \textbf{7.28} & \textbf{11.68} & \textbf{26.11} & \textbf{17.32} & \textbf{14.91} \\
\midrule
\multicolumn{15}{@{}l}{\textit{2 bits}} \\
GPTQ + clip & 2.141 & 10.49 & 8.31 & 39.51 & 129.9 & 32.87 & 21.74 & & 13.46 & 11.10 & 75.00 & 158.5 & 47.32 & 29.75 \\
GPTQ + clip, $G{=}64$ & 2.281 & 9.32 & 7.66 & 33.80 & \textbf{56.43} & 24.81 & 17.20 & & \textbf{12.02} & 10.30 & 68.22 & 78.43 & 35.07 & 24.33 \\
\weshareone{} & 2.156 & 9.25 & 7.43 & 16.35 & 63.63 & 20.61 & 15.44 & & 12.06 & 10.26 & 22.86 & 83.66 & 31.74 & 22.96 \\
\wesharetwo{} & 2.156 & \textbf{8.95} & \textbf{7.31} & \textbf{15.19} & 58.02 & \textbf{20.60} & \textbf{15.29} & & 12.07 & \textbf{10.11} & \textbf{21.83} & \textbf{77.45} & \textbf{31.14} & \textbf{22.58} \\
\bottomrule
\end{tabular*}
}
\end{table*}

\subsection{SliM-LLM and GPTAQ}
\label{sec:slim}

Table~\ref{tab:slim} compares with SliM-LLM and GPTAQ under SliM-LLM's protocol, including the 70B models. Figure \ref{fig:methods} and 
Table~\ref{tab:app-slim} (Appendix~\ref{app:results}) show these runs relative to FP16
and on three test sets. At 2 bits, \wesharetwo{} is below every other method we run on
all five models: 19-70\% below SliM-LLM, 12.5-35.3\% below GPTAQ and 2.5-79.1\% below GPTQ + clip. \weshareone{} is
also below SliM-LLM and GPTAQ on all five models. Both modes are below every method whose perplexity SliM-LLM reports,
including the three that learn their quantizer parameters by gradient descent: on Llama-2-13B and Llama-2-70B these reach
7.59 and 6.44 at best, against 6.70 and 4.71 for \wesharetwo{} and 6.88 and 4.76 for \weshareone{}. At 3 bits
\wesharetwo{} is 1.0-6.4\% below SliM-LLM on Llama-2 7B, 13B and 70B and Llama-3-8B, and 1.1\% above it on
Llama-3-70B, where it is level with GPTQ + clip at 4.58. Of the methods SliM-LLM reports, only SliM-LLM+ is below
\wesharetwo{} at 3 bits, on Llama-2-13B and Llama-2-70B (5.11 and 3.35 against 5.21 and 3.63). GPTAQ is level with
\wesharetwo{} on Llama-2-7B (5.93 against 5.92) and below it on Llama-2-13B (5.19 against 5.21). On the other
three models, \wesharetwo{} is 1.0-6.8\% below it. GPTQ + clip is itself below SliM-LLM in 7 of the 10 cells.

\wesharetwo{} is below \weshareone{} in all 10 cells of Table~\ref{tab:slim}. On Llama-3-70B at 2 bits, its per-group
layouts lower perplexity from 10.34 to 9.45.
At 2 bits a run also depends on floating-point rounding: over repeated runs on Llama-2-7B and Llama-3-8B
(Appendix~\ref{app:runs}), the variation is as large as the difference between the two modes, while \wesharetwo{} stays
below GPTQ + clip and both modes stay far below SliM-LLM in every run.

\subsection{Calibration on the Pile}
\label{sec:main}

Table~\ref{tab:main} compares both modes of \weshare{} with GPTQ at equal storage with calibration on the Pile, on
Llama-2, Llama-3 and Qwen2.5. Both modes are below GPTQ + clip on every model, at both widths and on both test sets. At
3 bits, \wesharetwo{} is below it by 0.7-9.7\% on WikiText-2 and 0.7-8.4\% on C4. The gain is largest on
Qwen2.5-0.5B and Llama-3-8B, the two models that GPTQ + clip degrades most. At 2 bits, where GPTQ + clip raises the
perplexity of FP16 by 70-92\% on Llama-2 and by factors of 6.4 and 9.9 on Llama-3-8B and Qwen2.5-0.5B, \wesharetwo{}
is 12-62\% below it on WikiText-2 and 9-71\% on C4. GPTQ + clip with groups of 64 stores 0.125 bits per weight more
and is still above \wesharetwo{} in 10 of the 12 cells.

In every cell, most of the gain over GPTQ + clip comes with \weshareone{} already. The per-group layouts of
\wesharetwo{} add to it at 2 bits: \wesharetwo{} is below \weshareone{} on every model on WikiText-2, by 3.6\% on
average and by up to 8.8\% on Qwen2.5-0.5B. On C4 it is below on five of the six models, by 2.8\% on average, and
level on Llama-2-7B (12.07 against 12.06). At 3 bits the average change is $-0.5\%$.

\subsection{Equal perplexity}
\label{sec:storage}

Layouts with $\delta<0$ trade bits for perplexity. Under SliM-LLM's protocol we evaluate
\weshareone{} with one layout in every group at U8 with $D\in\{16,48,96,120\}$ on the INT4 grid and
$D\in\{16,48,80,120\}$ on the INT3 grid, and at
$D=120$ with bit $b_1$ shared by four low-bin weights instead of two. The totals run from 3.164 down to 2.523 and
from 2.156 down to 1.516 bits per weight.
Table~\ref{tab:iso} gives the storage at which this curve, interpolated log-linearly in perplexity, reaches the perplexity of SliM-LLM's code,
and Figure~\ref{fig:curves} in Appendix~\ref{app:collapse} shows the curves.
At 2 bits, \weshareone{} needs 17.9-19.4\% fewer bits per weight on WikiText-2, 1.726-1.758 instead of 2.141, and
21.6\% or more on C4. At 3 bits the saving is 2.1-14.7\%. On the 70B models, where SliM-LLM's code searches its own
widths, the saving at 2 bits is 18.2\% on Llama-2-70B and 13.6\% on Llama-3-70B, and at 3 bits 1.9\% on Llama-2-70B.
On Llama-3-70B, \weshareone{} with U8~D16 (3.164 bits per weight) is above SliM-LLM's 3-bit perplexity. The
curve continues upward with U8~D0 and U32~D0 (3.227 and 3.414) and reaches it at 3.200, 1.7\% more than SliM-LLM. These
savings are those of \weshareone{}. On WikiText-2, \wesharetwo{} is lower still at 3.164 and 2.156 bits per weight.

\begin{figure*}[!ht]
  \centering
  \subfigure[Layer Decoder.]{
        \label{fig:decoder}
        \includegraphics[width=0.48\textwidth]{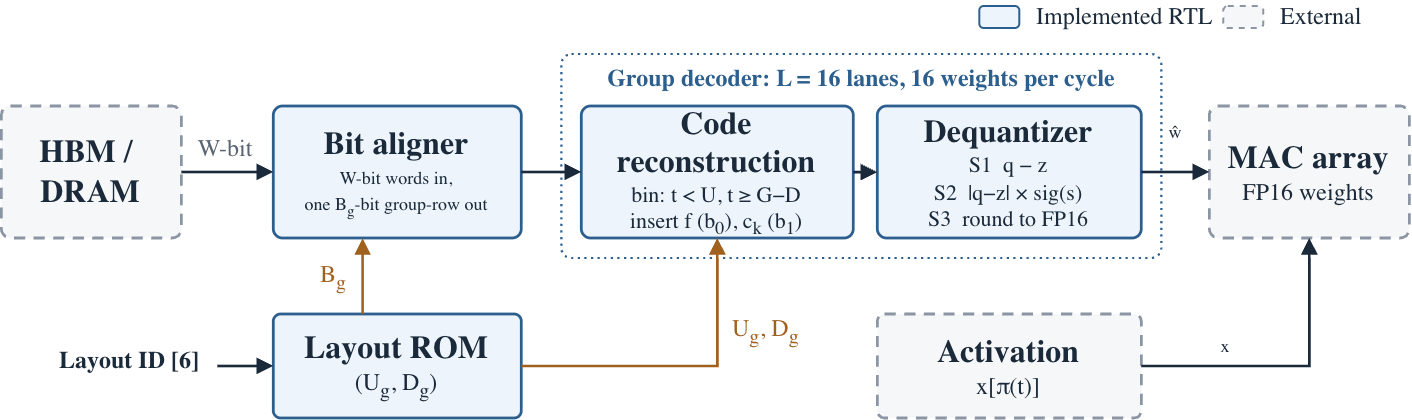}}
    \subfigure[Code reconstruction.]{
        \label{fig:reconst}
        \includegraphics[width=0.27\textwidth]{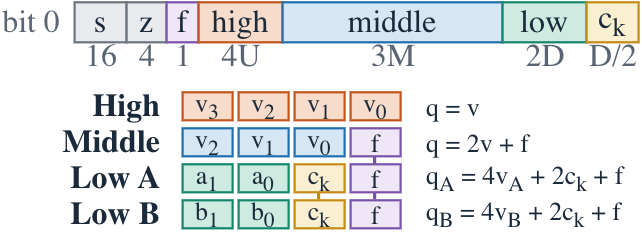}}
    \subfigure[One group-row.]{
        \label{fig:group}
        \includegraphics[width=0.21\textwidth]{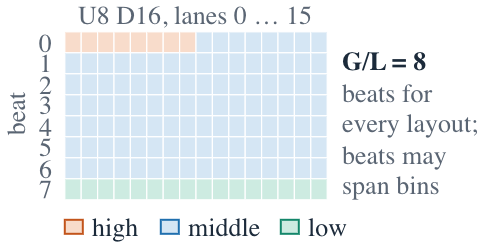}}
  \caption{\weshare{} decoder hardware implementation. (a)~Each group's layout index selects $(U_g,D_g)$ and the group-row size
$B_g$; the bit aligner extracts each group-row from the packed layer, and the
group decoder rebuilds and dequantizes its codes. Activations are read through
$\pi$. (b)~Code reconstruction: $f$ supplies $b_0$ of
middle- and low-bin codes, and $c_k$ supplies $b_1$ of both codes of a low-bin
pair. (c)~Outputs may span bins, so every layout takes $G/L$ outputs of $L$
weights per group-row.}
  \label{fig:bitshare-decoder}
\end{figure*}

\begin{table}[!ht]
\caption{Storage saved at equal perplexity: the reduction in bits per weight against SliM-LLM (3.148 and 2.141) at which the rate-quality curve of \weshareone{}, with one layout in every group, reaches the perplexity of SliM-LLM's code, both under SliM-LLM's protocol. On the 70B models, SliM-LLM's code searches its own widths, and only WikiText-2 is evaluated. $\ge$: SliM-LLM's perplexity is above this curve, which ends at 1.516 bits per weight; negative: more bits than SliM-LLM.}
\label{tab:iso}
\centering
\footnotesize
\begin{tabular}{@{}l cc cc@{}}
\toprule
 & \multicolumn{2}{c}{3 bits} & \multicolumn{2}{c}{2 bits} \\
\cmidrule(lr){2-3}\cmidrule(l){4-5}
Model & WikiText-2 & C4 & WikiText-2 & C4 \\
\midrule
Llama-2-7B & 14.7\% & 5.7\% & 19.4\% & 28.5\% \\
Llama-2-13B & 2.1\% & 3.2\% & 18.5\% & 21.6\% \\
Llama-3-8B & 5.2\% & 10.7\% & 17.9\% & $\ge$29.2\% \\
Llama-2-70B & 1.9\% & -- & 18.2\% & -- \\
Llama-3-70B & $-$1.7\% & -- & 13.6\% & -- \\
\bottomrule
\end{tabular}
\end{table}

\subsection{Down to collapse}
\label{sec:collapse}

We extend \weshareone{} below 1.516 bits per weight on the INT3 grid under SliM-LLM’s protocol using U0 D128 with $b_1$ shared by 4/8/16 weights or dropped (1.406/1.281/1.219/1.156 bits per weight), U8 D120 with eight-way sharing or dropping (1.398/1.281) and U0 D256 with 256-weight groups and $b_1$ dropped (1.078).

Table~\ref{tab:collapse} in Appendix~\ref{app:collapse} gives the rate at which WikiText-2 perplexity, with the best of these formats at each rate,
reaches two and ten times that of FP16. In Llama-2 a larger model holds out longer: perplexity doubles at 2.02, 1.72
and 1.61 bits per weight and reaches ten times FP16 at 1.36, 1.22 and 1.17 from 7B to 70B. Llama-3-70B doubles it
already at 2.91 bits per weight, against 2.30 for Llama-3-8B, but then degrades more slowly and reaches ten times FP16
at 1.41, against 1.68. Sharing $b_1$ keeps a model usable longer than dropping it: at 1.281 bits per weight,
perplexity with $b_1$ shared by eight weights is 2.1, 2.7 and 1.7 times lower than without $b_1$ on Llama-2 7B, 13B
and 70B, and 32 and 1.2 times lower on Llama-3 8B and 70B. Up to three times FP16 each curve follows
$\ln d=a_m+\beta_m b$, with $d=\ln(\mathrm{ppl}/\mathrm{ppl}_{\mathrm{FP16}})$. With the slope taken from the other models, a
single run of a model predicts the rest of its curve with a median error of 5.2\% in perplexity
(Appendix~\ref{app:collapse}).

\subsection{Ablations}
\label{sec:ablation}

Table~\ref{tab:ablation} in Appendix~\ref{app:ablation} starts from GPTQ + clip on the INT2 grid and isolates the
parts of the 2-bit format one at a time. Without the bins (U0~D0), perplexity is within 0.6\% of \weshareone{} on Llama-2 but 8-65\% higher on
Llama-3-8B and the Qwen models, where a few channels per group need the full INT3 grid of the high bin. Taking bit $b_0$
from the parity of the zero point instead of the flag saves 1/128 bit per weight but raises perplexity by 5-30\%. A
layout per group with equal bits in every group lowers perplexity against \weshareone{} by 2.6\% on average, though not
on every model, and moving bits between groups by Equation~\ref{eq:alloc}, as \wesharetwo{} does, lowers it on every model, by
3.6\% on average. The appendix also gives the details and shows the layouts that \wesharetwo{} chooses
(Figure~\ref{fig:bins}).

\subsection{Encoding \texorpdfstring{and decoding}{and decoding} cost}
\label{sec:cost}

\paragraph{Encoding} \wesharetwo{} chooses the layouts from the weights and the diagonal of the Hessian that GPTQ accumulates anyway,
and both modes find the codes in one GPTQ pass. On one 48\,GB GPU, quantizing Llama-2-70B or
Llama-3-70B takes 1.8-1.9 hours with GPTQ + clip, 2.2-2.3 hours with \weshareone{} and 3.1 hours with \wesharetwo{},
at either width. SliM-LLM scores every candidate allocation of a linear layer by the KL divergence of its outputs on the
calibration set. Its code spends 32-38 minutes per decoder layer on this
search alone, or 43-51 hours per model and width, also on one GPU.

\paragraph{Decoding} 
Assuming weights are streamed once per token at batch size one, bits per weight constrain bandwidth-limited decoding throughput.
Both modes use 3.164/2.156 bits per weight in group records, approximately 0.016 above INT3/INT2 and SliM-LLM. 
At SliM-LLM's 2-bit perplexity on WikiText-2, however, \weshareone{} reads 13.6-19.4\% fewer bits (Table~\ref{tab:iso}),
and Appendix~\ref{app:hw-bound} turns these savings into a bound on the speedup. Neither mode has per-weight metadata or a
codebook. In \weshareone{} every group has the same layout, so each field sits at a fixed offset and no layout table
is needed. \wesharetwo{} reads a 6-bit layout index per group and selects each field at a layout-dependent offset.
Section~\ref{sec:hw_impl} implements its decoder in hardware. SliM-LLM, whose groups also differ in size, reports
20-27\% lower GPU throughput than GPTQ at 2 bits with its own kernels, on three models and an A800 GPU (Table~4
of~\cite{huang2025slimllm}).

\section{Hardware Implementation}
\label{sec:hw_impl}

We implemented
the decoder of \weshare{} in synthesizable SystemVerilog code, verified it and compared its resource overhead with
INT-$n$ and SliM-LLM under similar constraints. Figure~\ref{fig:bitshare-decoder} shows the data path from the
compressed stream to FP16 weights (Figure \ref{fig:decoder}) and
the reconstruction from stored and shared bits (Figure \ref{fig:reconst}) along with the
layout-independent output schedule of $G/L$ outputs per group-row (Figure \ref{fig:group}).

\subsection{Microarchitecture}
\label{sec:hw-arch}

In the decoder implementation, group-rows are stored back to
back without padding, and one 6-bit layout index per group, shared by all
rows of a layer, gives each group-row its layout $(U_g,D_g)$ and size $B_g$.
A bit aligner cuts each group-row from the packed stream, and the group
decoder rebuilds the $n$-bit codes by Equation~\ref{eq:codes} and dequantizes them to
binary16 by Equation~\ref{eq:dequant}, exactly as for INT-$n$. It emits $L$ weights per cycle
and takes $G/L$ cycles per group-row for every layout, so for $G/L\ge2$ it
sustains $\min(L,\,WG/\bar B)$ weights per cycle from a $W$-bit memory
interface, where $\bar B$ is the mean group-row size; at $G=128$ and $L=16$,
a 64-bit interface keeps up with every layout. Appendix~\ref{app:hw-arch}
includes the detailed implementation.

\subsection{Hardware cost}
\label{sec:hw-cost}

\subsubsection{Baselines}  We compare three decoders with identical interfaces, input buffering, three-stage pipelines and binary16 dequantizers, each delivering 16 weights per cycle at a group size of 128: INT-$n$ (GPTQ storage), with fixed $n$-bit fields and lane offsets; SliM-LLM, with $n-1$, $n$ or $n+1$ bits per weight selected per group by a 2-bit code shared across rows, densely packed records and affine reconstruction; and \weshare{} with per-group layouts and two-way sharing. SliM-LLM and \weshare{} require bit alignment for variable-size records. We also equip INT-$n$ with the same aligner, although its fixed-size records permit a simpler design. Area is the sum of the mapped cell areas, without placement, routing or clock tree. 
Logic delay is the critical combinational path reported by ABC.

\subsubsection{Setup}  We flatten and synthesize each design using open-source Yosys and ABC \cite{wolf2013yosys,brayton2010abc} tools with the SkyWater 130\,nm (sky130\_fd\_sc\_hd) \cite{skywater2020pdk} library at the typical corner (25°C, 1.80 V). Area sums mapped standard-cell areas, excluding placement, routing and clock-tree overhead. The reports include critical combinational delay, excluding register, setup and interconnect delays.

\begin{table}[!ht]
\centering
\caption{Decoder area at equal throughput (16 weights/cycle, $G=128$),
SkyWater 130\,nm. \emph{Decoder}: the group decoder plus the bit aligner and, for \weshare{}, the layout ROM. Flip-flops and logic delay are calculated for the group decoder.}
\label{tab:hw-area}
\small
\resizebox{\columnwidth}{!}{%
\begin{tabular}{@{}llrrrrr@{}}
\toprule
 & & \multicolumn{2}{c}{Area ($\mu$m$^2$)} & & & Logic \\
\cmidrule(lr){3-4}
Rate & Format & Group dec. & Decoder & vs.\ INT-$n$ & Flip-flops & delay (ns) \\
\midrule
3 bits & INT3     & 100{,}693 & 183{,}286 & 1.00$\times$ & 1{,}795 & 6.39 \\
       & SliM-LLM & 129{,}728 & 237{,}218 & 1.29$\times$ & 2{,}169 & 7.36 \\
       & \weshare & 210{,}243 & 318{,}134 & 1.74$\times$ & 2{,}327 & 7.88 \\
\midrule
2 bits & INT2     &  80{,}106 & 139{,}928 & 1.00$\times$ & 1{,}425 & 6.15 \\
       & SliM-LLM & 105{,}057 & 188{,}672 & 1.35$\times$ & 1{,}799 & 6.13 \\
       & \weshare & 175{,}384 & 258{,}311 & 1.85$\times$ & 1{,}957 & 7.63 \\
\bottomrule
\end{tabular}
}
\end{table}

\subsubsection{Results} Table~\ref{tab:hw-area} presents the
hardware overhead for the \weshare{} decoder implementation. At equal throughput, the \weshare{} decoder takes 1.34$\times$ the
area of the SliM-LLM decoder at 3 bits and 1.37$\times$ at 2 bits, and
1.74$\times$ and 1.85$\times$ that of INT-$n$. Its logic delay is
1.23$\times$ and 1.24$\times$ that of INT-$n$, and 1.07$\times$ and
1.24$\times$ that of SliM-LLM.

At 3 bits, the
three group decoders incur 53.8 (INT3), 65.0 (SliM-LLM) and
69.8\,k$\mu$m$^2$ (\weshare) of flip-flop area, but 46.9, 64.7 and
140.5\,k$\mu$m$^2$ of logic area. An INT-$n$ lane reads a fixed bit slice and a
SliM-LLM lane chooses among three fixed strides. 
A \weshare{} lane selects
its field at a layout-dependent offset, and the active group advances by a
layout-dependent number of bits per beat. All three share the same binary16 dequantizer, while SliM-LLM and \weshare{} have nearly identical bit-aligner area.

This overhead supports a different quality-storage trade-off. At comparable storage, \weshare{} adds approximately 0.016 bits per weight and achieves lower perplexity in most evaluated settings. At matched perplexity, the configurations reduce storage (Table~\ref{tab:iso}) and weight traffic (Appendix \ref{app:hw-bound}) in most cases. The decoding circuitry can be reused across groups and layers, so its area scales with the number of instantiated decoding engines rather than the number of stored weights. Consequently, the decoder-only area ratios do not represent whole-accelerator overhead: the latter also depends on the area of the compute array, memories and interconnect, together with the required decoder replication.

\subsubsection{Energy savings} We consider batch-1 decoding in which quantized weights are streamed from DRAM once per generated token. At equal perplexity, fewer bits per weight reduce storage and, consequently, the weight traffic. Under a constant energy-per-transferred-bit model, this yields proportional savings in weight-transfer energy (Appendix~\ref{app:hw-energy}).

\section{Conclusion}
\label{sec:conclusion}

We have presented \weshare{}, a group-wise weight format that combines sensitivity-aware binning and static bit sharing with optimal bit-value allocation to improve the quality-storage trade-off of LLM weight quantization without codebooks or fine-tuning.
With calibration on the Pile and at the same storage, both modes of \weshare{} have lower perplexity than GPTQ with the same range search, and at 2 bits \wesharetwo{} lowers WikiText-2 perplexity by up to 70\% against SliM-LLM and up to 35\% against GPTAQ. At SliM-LLM's 2-bit
perplexity, \weshareone{} stores up to 19.4\% fewer bits per weight. The hardware implementation of the decoder verifies correct operation and practical feasibility while characterizing its modest overhead. Future work will include measuring end-to-end decoding speed on GPUs, power analysis and combining bit sharing with gradient-based calibration.

\bibliographystyle{IEEEtran}
\bibliography{refs}
\clearpage
\appendices

\section{Algorithm}
\label{app:algo}

{Algorithm~\ref{alg:encoder} gives the encoder of \weshare{} for one linear layer with weights
$\mathbf{W}\in\mathbb{R}^{R\times C}$ and the Hessian $\mathbf{H}$ accumulated from its calibration inputs. It returns the
layout indices, the stored fields, the pair bits, the group parameters and the channel permutation $\pi$.}

\begin{algorithm}[!ht]
\caption{\weshare{} encoder for one linear layer. {\weshareone{} runs it with a one-layout menu, so lines 2-5 give
every group that layout.}}
\label{alg:encoder}
\begin{algorithmic}[1]
\REQUIRE weights $\mathbf{W}\in\mathbb{R}^{R\times C}$, Hessian $\mathbf{H}$ from calibration inputs, grid width $n$,
         layout menu $\Lambda$, clipping factors $P$
\STATE $h_j\leftarrow[\mathbf{H}]_{jj}$
\FOR{each group $g$ and layout $\ell\in\Lambda$}
  \STATE $E_g(\ell)\leftarrow\sum_i\min_{(p,f)\in P\times\{0,1\}}\sum_{j\in g}h_j(w_{ij}-\hat w^{(\ell)}_{ij})^2$ under
         round-to-format codes
\ENDFOR
\STATE choose $(\ell_g)_g$ by Equation~\ref{eq:alloc} (dynamic programming over $\sum_g\delta(\ell_g)$)
\STATE build $\pi$ from $h$ and the layouts (Section~\ref{sec:binning}); permute the columns of $\mathbf{W}$ and the rows and
       columns of $\mathbf{H}$ by $\pi$
\FOR{each group-row $(i,g)$}
  \STATE $(s_{ig},z_{ig},f_{ig})\leftarrow$ the $(p,f)\in P\times\{0,1\}$ with the smallest
         $\sum_j h_j(w_{ij}-\hat w_{ij})^2$ under round-to-format codes of layout $\ell_g$
\ENDFOR
\STATE units $\leftarrow$ low-bin pairs (blocks of $B$) and all other single columns, by decreasing
       $\max_{j\in\text{unit}}h_j$
\STATE $\mathbf{L}\leftarrow$ upper Cholesky factor of $(\mathbf{H}+\lambda\mathbf{I})^{-1}$ in unit order,
       $\lambda=0.01\,\overline{h}$; $d_j\leftarrow L_{jj}$
\FOR{each unit $u$ in order}
  \IF{$u$ is a column $j$}
    \STATE $q_{ij}\leftarrow$ nearest code allowed in $j$'s bin, for every row $i$
  \ELSE
    \FOR{$c\in\{0,1\}$}
      \STATE round the unit's columns in turn with $b_1=c$, each propagating its error to the next
      \STATE $J_i(c)\leftarrow\sum_{j\in u}(w_{ij}-\hat w_{ij})^2/d_j^2$ for every row $i$
    \ENDFOR
    \STATE $c_{igk}\leftarrow\arg\min_c J_i(c)$ and keep its codes, for every row $i$
  \ENDIF
  \STATE propagate the unit's errors $(w_{ij}-\hat w_{ij})/d_j$ to the remaining columns through $\mathbf{L}$
\ENDFOR
\STATE \textbf{return} layout indices, fields, pair bits, $s$, $z$, $f$ and $\pi$
\end{algorithmic}
\end{algorithm}

\section{\texorpdfstring{{Format and encoder details}}{Format and encoder details}}
\label{app:format}

\noindent\textbf{The flag.} The flag fixes a bias of the narrower grids. The $2^{n-1}$ codes of one parity cover the group's range except one fine
step at one end: the codes of parity 0 miss the top code $2^n-1$, those of parity 1 miss code $0$. If $b_0$ were tied to
a value the decoder already holds, such as the parity of the zero point, the uncovered end would follow the sign balance
of the group, through $z_{ig}$, rather than where its important weights lie. The flag lets the encoder choose that end,
and with it the offset of the low bin's grid, for one bit per group-row. Equivalently, the flag is a bit shared by all
$M_g+D_g$ middle- and low-bin weights of the group-row. On the INT4 grid the middle bin is then a 3-bit grid whose step,
$2/15$ of the range, is 6.7\% finer than INT3's $1/7$; on the INT3 grid it is a 2-bit grid with step $2/7$, 14\% finer
than INT2's $1/3$.

\noindent\textbf{Lower rates.} Layouts with $\delta<0$ store fewer bits. To go further, the pair bit generalizes to a bit
shared by $B$ consecutive low-bin positions, {which gives} $D/B$ bits per group-row and $\delta=U-D(1-1/B)$.
{Without such a bit, $b_1$ is \emph{dropped}: every low-bin code takes it from bit 1 of the zero point $z_{ig}$, and
$\delta=U-D$.} {Both modes use pairs ($B=2$). To extend the rate-quality curve of \weshareone{} down to collapse,
Sections~\ref{sec:storage} and~\ref{sec:collapse} also use $B\in\{4,8,16\}$ and the dropped bit in every group.}

\begin{figure}[t]
\centering
\begin{tikzpicture}[x=0.17mm,y=4.2mm,font=\scriptsize]
  \foreach \g/\u/\d/\name/\off in {0/32/0/{U32 D0}/{$\delta=+32$}, 1/8/16/{U8 D16}/{$\delta=0$}, 2/0/64/{U0 D64}/{$\delta=-32$}} {
    \pgfmathsetmacro{\xo}{\g*142}
    \pgfmathsetmacro{\m}{128-\u-\d}
    \pgfmathsetmacro{\xl}{\xo+128-\d}
    \ifdim \u pt>0pt \draw[fill=wsHigh,draw=wsLine] (\xo,0) rectangle ++(\u,1); \fi
    \draw[fill=wsMid,draw=wsLine] (\xo+\u,0) rectangle ++(\m,1);
    \ifdim \d pt>0pt \draw[fill=wsLow,draw=wsLine] (\xl,0) rectangle ++(\d,1); \fi
    \node[above] at (\xo+64,1) {\name};
    \node[below] at (\xo+64,0) {\off};
  }
  \draw[fill=wsHigh,draw=wsLine] (0,-2.2) rectangle ++(9,0.6);
  \node[anchor=west] at (11,-1.9) {high, $n$ bits};
  \draw[fill=wsMid,draw=wsLine] (140,-2.2) rectangle ++(9,0.6);
  \node[anchor=west] at (151,-1.9) {middle, $n{-}1$};
  \draw[fill=wsLow,draw=wsLine] (282,-2.2) rectangle ++(9,0.6);
  \node[anchor=west] at (293,-1.9) {low, $n{-}2$ + pairs};
\end{tikzpicture}
\caption{Three groups of one row {in \wesharetwo{},} with different layouts, $G=128$ positions each{. In \weshareone{}
every group has the same layout}. Bits per group-row differ from the reference by $\delta=U-D/2${, and} within a layer the
offsets sum to at most zero. Every row of the layer has the same layouts.}
\label{fig:layouts}
\end{figure}
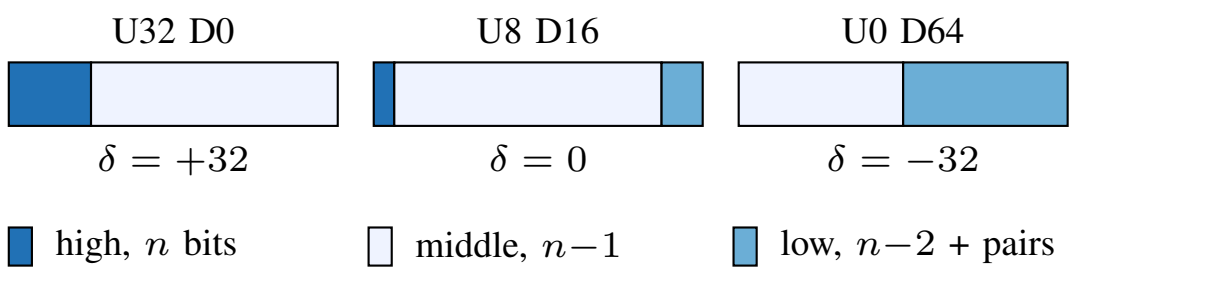

\noindent\textbf{Layouts.} {Figure~\ref{fig:layouts} shows three layouts of the menu of \wesharetwo{}.} The layouts follow the groups' sensitivity without being set by it alone: the rank correlation between a group's mean $h_j$ and its offset is $+0.3$ to $+0.7$ across the models {we evaluate}.

\noindent\textbf{Rounding.} {For a bit shared by $B$ low-bin positions, a block of $B$ columns is rounded as a pair is, one
column after another.} A weight is always rounded to the nearest code consistent with the
shared bits, which may change its stored field as well (a carry or borrow), rather than having its own shared bit
overwritten.

\noindent\textbf{Extra work of the encoder.} The encoder needs no fine-tuning and no codebook. Its extra work over GPTQ with the same range search is the range
search for each layout of the menu before the column loop {(in \wesharetwo{})}, the second flag value in that search and
the two candidate roundings per low-bin pair and row.

\section{Additional results}
\label{app:results}

Table~\ref{tab:app-w4} repeats the setting of Table~\ref{tab:main} on WikiText-2 with windows of 4096 tokens.
Table~\ref{tab:app-slim} compares {both modes of} \weshare{} with our runs of SliM-LLM's and GPTAQ's code on all three test
sets, under SliM-LLM's protocol.

On all three test sets of Table~\ref{tab:app-slim}, {\wesharetwo{}} is {below SliM-LLM by 29-93\% at 2 bits and
by 0.9-10.5\% at 3 bits. It is also 14-98\% below GPTAQ at 2 bits.} At 3 bits it is 2.5-5.5\% below GPTAQ on C4
and 0.9-3.8\% below it on WikiText-2 for Llama-3-8B, while on WikiText-2 for Llama-2 GPTAQ is lower in three of the
four cells, by at most 0.4\%.

\begin{table}[!ht]
\caption{WikiText-2 perplexity with 4096-token windows, calibration on the Pile (the setting of Table~\ref{tab:main}). The lowest perplexity of each width is in \textbf{bold}.}
\label{tab:app-w4}
\centering
\footnotesize
\setlength{\tabcolsep}{2.6pt}
\begin{tabular}{@{}lc rr r rrr@{}}
\toprule
 & & \multicolumn{2}{c}{Llama-2} & Llama-3 & \multicolumn{3}{c}{Qwen2.5} \\
\cmidrule(lr){3-4}\cmidrule(lr){5-5}\cmidrule(l){6-8}
Method & b/w & 7B & 13B & 8B & 0.5B & 1.5B & 3B \\
\midrule
FP16 & 16 & 5.12 & 4.57 & 5.75 & 12.08 & 8.57 & 7.45 \\
\midrule
\multicolumn{8}{@{}l}{\textit{3 bits}} \\
GPTQ & 3.148 & 5.82 & 5.05 & 7.42 & 18.06 & 11.15 & 9.22 \\
GPTQ + clip & 3.148 & 5.67 & 5.10 & 7.32 & 17.60 & 10.40 & 8.76 \\
\weshareone{} & 3.164 & \textbf{5.62} & 5.00 & 6.95 & 16.18 & \textbf{10.10} & 8.60 \\
\wesharetwo{} & 3.164 & 5.63 & \textbf{4.95} & \textbf{6.93} & \textbf{15.92} & 10.11 & \textbf{8.56} \\
\midrule
\multicolumn{8}{@{}l}{\textit{2 bits}} \\
GPTQ + clip & 2.141 & 9.83 & 7.75 & 39.38 & 109.7 & 30.12 & 19.75 \\
\quad $G{=}64$ & 2.281 & 8.72 & 7.15 & 33.80 & \textbf{50.91} & 22.85 & 15.77 \\
\weshareone{} & 2.156 & 8.62 & 6.98 & 16.01 & 57.84 & \textbf{18.90} & 14.19 \\
\wesharetwo{} & 2.156 & \textbf{8.38} & \textbf{6.79} & \textbf{13.75} & 53.83 & 18.96 & \textbf{14.09} \\
\bottomrule
\end{tabular}
\end{table}

\begin{table*}[!ht]
\caption{Perplexity under SliM-LLM's protocol (calibration on WikiText-2) against our runs of SliM-LLM's code with its released group widths and of GPTAQ's code. W2: WikiText-2 with 2048- or 4096-token windows. The lowest perplexity of each width is in \textbf{bold}.}
\label{tab:app-slim}
\centering
\footnotesize
\setlength{\tabcolsep}{4pt}
\begin{tabular}{@{}lc rrr c rrr c rrr@{}}
\toprule
 & & \multicolumn{3}{c}{Llama-2-7B} & & \multicolumn{3}{c}{Llama-2-13B} & & \multicolumn{3}{c}{Llama-3-8B} \\
\cmidrule(lr){3-5}\cmidrule(lr){7-9}\cmidrule(l){11-13}
Method & b/w & W2 2048 & W2 4096 & C4 & & W2 2048 & W2 4096 & C4 & & W2 2048 & W2 4096 & C4 \\
\midrule
FP16 & 16 & 5.47 & 5.12 & 7.26 & & 4.88 & 4.57 & 6.73 & & 6.14 & 5.75 & 9.44 \\
\midrule
\multicolumn{13}{@{}l}{\textit{3 bits}} \\
SliM-LLM (code) & 3.148 & 6.33 & 5.85 & 8.55 & & 5.26 & 4.92 & 7.50 & & 7.69 & 7.16 & 13.74 \\
GPTAQ (code) & 3.148 & 5.93 & \textbf{5.53} & 8.70 & & \textbf{5.19} & \textbf{4.86} & 7.55 & & 7.37 & 7.11 & 12.74 \\
GPTQ + clip & 3.148 & 5.96 & 5.56 & 8.30 & & 5.33 & 4.99 & 7.58 & & 7.34 & 6.88 & 12.74 \\
\weshareone{} & 3.164 & 5.94 & 5.54 & \textbf{8.22} & & 5.22 & 4.89 & 7.38 & & 7.42 & 6.94 & 12.38 \\
\wesharetwo{} & 3.164 & \textbf{5.92} & 5.54 & \textbf{8.22} & & 5.21 & 4.87 & \textbf{7.36} & & \textbf{7.31} & \textbf{6.85} & \textbf{12.29} \\
\midrule
\multicolumn{13}{@{}l}{\textit{2 bits}} \\
SliM-LLM (code) & 2.141 & 17.31 & 19.67 & 64.27 & & 9.47 & 8.78 & 18.79 & & 41.83 & 44.35 & 434.7 \\
GPTAQ (code) & 2.141 & 11.76 & 13.64 & 28.93 & & 7.84 & 7.46 & 16.57 & & 20.87 & 18.76 & 1.2e3 \\
GPTQ + clip & 2.141 & 10.37 & 9.65 & \textbf{17.79} & & 7.41 & 6.84 & 12.02 & & 21.95 & 21.22 & 228.6 \\
\weshareone{} & 2.156 & 10.53 & 9.75 & 18.27 & & 6.88 & 6.40 & 10.81 & & 15.06 & 13.87 & 33.65 \\
\wesharetwo{} & 2.156 & \textbf{10.11} & \textbf{9.25} & 18.72 & & \textbf{6.70} & \textbf{6.25} & \textbf{10.46} & & \textbf{14.47} & \textbf{13.29} & \textbf{30.81} \\
\bottomrule
\end{tabular}
\end{table*}

\subsection{\texorpdfstring{{Baseline runs}}{Baseline runs}}
\label{app:baselines}

\noindent\textbf{SliM-LLM.} {On one linear layer of Llama-2-70B, its salience computation fails in the Cholesky factorization, which we repeat with more damping.} With the released widths our runs are within 1.4\% of the perplexities SliM-LLM reports at 3 bits and 0.6-11.8\% above them at 2 bits. {On the 70B models they range from 6.9\% below to 37\% above them.}

\noindent\textbf{GPTAQ.} At 3 bits we run it with the groups fixed before quantization and with the groups formed in act-order and report the better run for each model. {At 2 bits, where the second diverges on Llama-2-7B, the groups are fixed.}

\subsection{Repeated runs}
\label{app:runs}

At 2 bits the result of a run depends on floating-point rounding. On an RTX 6000 Ada GPU
instead of an RTX A6000, \weshareone{} gives 8.78 instead of 10.53 on Llama-2-7B. {The two GPUs round the outputs of
the first decoder layer differently. The Hessians of its query projection then differ by $8\cdot10^{-7}$ relative, and
GPTQ already gives 0.16\% of its weights different 2-bit codes.} On one GPU type a run is deterministic. Table~\ref{tab:app-runs}
repeats the 2-bit runs of GPTQ + clip, \weshareone{} and \wesharetwo{} on Llama-2-7B and Llama-3-8B, twice with the
calibration sequences in another order, which changes only the rounding of the sums over them, and twice with other
sequences. Over the five runs, the standard deviation of the logarithm of perplexity is 0.05-0.06 for \weshareone{} and
\wesharetwo{}, and 0.03 and 0.22 for GPTQ + clip on Llama-2-7B and Llama-3-8B. \wesharetwo{} is below GPTQ + clip in all ten
runs, by 8.4\% and 42\% on average, and every run {of both modes} stays far below SliM-LLM's code (17.31 and 44.35).
{Against \weshareone{}, \wesharetwo{}} is lower in three of the five runs on Llama-2-7B and in two on Llama-3-8B{.
On average} it is 1.3\% lower on Llama-2-7B and 1.0\% higher on Llama-3-8B. On these two models, the {gain of \wesharetwo{}
over \weshareone{}} at 2 bits under SliM-LLM's protocol is within the variation between runs.

\begin{table}[t]
\caption{Repeated runs at 2 bits under SliM-LLM's protocol: WikiText-2 perplexity of the run of Table~\ref{tab:slim} (run), of the same run with the 128 calibration sequences in two other orders {(order) and} with two other draws of the sequences (seed). {A new order changes only the rounding of the sums over the sequences.} Windows of 2048 tokens on Llama-2-7B and 4096 on Llama-3-8B, as in Table~\ref{tab:slim}. Mean: geometric mean; sd: standard deviation of the logarithm, over the five runs.}
\label{tab:app-runs}
\centering
\footnotesize
\setlength{\tabcolsep}{2pt}
\begin{tabular}{@{}l rrrrr rr@{}}
\toprule
Method & run & order 1 & order 2 & seed 1 & seed 2 & mean & sd \\
\midrule
\multicolumn{8}{@{}l}{\textit{Llama-2-7B}} \\
GPTQ + clip & 10.37 & 10.40 & 10.43 & 9.86 & 9.87 & 10.18 & 0.029 \\
\weshareone{} & 10.53 & 9.40 & 9.18 & 8.91 & 9.34 & 9.46 & 0.064 \\
\wesharetwo{} & 10.11 & 9.14 & 9.24 & 9.04 & 9.16 & 9.33 & 0.046 \\
\midrule
\multicolumn{8}{@{}l}{\textit{Llama-3-8B}} \\
GPTQ + clip & 21.22 & 20.89 & 26.37 & 22.72 & 35.64 & 24.84 & 0.222 \\
\weshareone{} & 13.87 & 14.98 & 14.85 & 14.02 & 13.12 & 14.15 & 0.054 \\
\wesharetwo{} & 13.29 & 15.03 & 13.85 & 14.76 & 14.59 & 14.29 & 0.051 \\
\bottomrule
\end{tabular}
\end{table}

\subsection{\texorpdfstring{{Down to collapse}}{Down to collapse}}
\label{app:collapse}

\begin{figure}[!ht]
\centering
\includegraphics[width=\columnwidth]{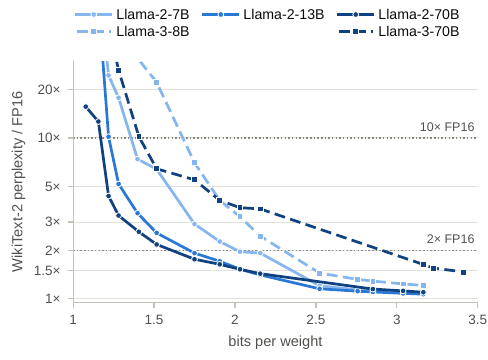}
\caption{{Rate-quality curves of \weshareone{} under SliM-LLM's protocol: WikiText-2 perplexity (2048-token
windows) relative to FP16 against bits per weight, with the best format at each rate. Lighter lines: smaller models;
dashed: Llama-3. Dotted: 2 and 10 times FP16 (Table~\ref{tab:collapse}).}}
\label{fig:curves}
\end{figure}

\noindent\textbf{A law for the curve.} With $d=\ln(\mathrm{ppl}/\mathrm{ppl}_{\mathrm{FP16}})$, each curve follows
$\ln d=a_m+\beta_m b$ up to three times FP16, where $b$ is the bits per weight. The slopes (Table~\ref{tab:collapse})
flatten with model size within each family, and one common slope, $\beta=-1.59$, with one intercept per model fits all
37 points with an error of 0.17 in $\ln d$. A single run of a model therefore predicts its curve: fixing $a_m$ with the
run nearest 2.156 bits per weight below three times FP16 and taking $\beta$ from the other models predicts the other 32
points with a median error of 5.2\% in perplexity, the size of the run-to-run variation at 2 bits (a standard deviation
of 5-7\%, Appendix~\ref{app:runs}). {The largest error, 48\%, is on Llama-2-70B, the Llama-2 model with the flattest slope, at 1.406 bits per
weight. The law holds only up to three times FP16. Before we ran the 70B models below 1.516 bits per weight,
extrapolating it placed ten times FP16 at 0.6-0.9 bits per weight on Llama-2-70B, which reaches it at 1.17, and at
1.2-1.6 on Llama-3-70B, which reaches it at 1.41.}

\begin{table}[!ht]
\caption{Down to collapse, {\weshareone{} (one layout in every group)} under SliM-LLM's protocol: the rate (bits per weight) at which WikiText-2 perplexity, with the best format evaluated at each rate, reaches 2 and 10 times that of FP16{. $\beta_m$: the slope of $\ln d$ against bits per weight up to 3 times FP16, with $d=\ln(\mathrm{ppl}/\mathrm{ppl}_{\mathrm{FP16}})$. 1.281 b/w: the perplexity at 1.281 bits per weight} with bit $b_1$ shared by eight low-bin weights (U0~D128) and with $b_1$ dropped (U8~D120).}
\label{tab:collapse}
\centering
\footnotesize
\setlength{\tabcolsep}{4pt}
\begin{tabular}{@{}l cc r rr@{}}
\toprule
 & \multicolumn{2}{c}{$\times$ FP16} & & \multicolumn{2}{c}{1.281 b/w} \\
\cmidrule(lr){2-3}\cmidrule(l){5-6}
Model & 2 & 10 & $\beta_m$ & shared & dropped \\
\midrule
Llama-2-7B & 2.02 & 1.36 & $-$1.97 & 97.30 & 199.8 \\
Llama-2-13B & 1.72 & 1.22 & $-$1.69 & 25.28 & 67.50 \\
Llama-2-70B & 1.61 & 1.17 & $-$1.30 & 10.93 & 18.94 \\
Llama-3-8B & 2.30 & 1.68 & $-$1.50 & 904.7 & 2.9e4 \\
Llama-3-70B & 2.91 & 1.41 & $-$0.93 & 75.27 & 87.33 \\
\bottomrule
\end{tabular}
\end{table}

\subsection{\texorpdfstring{{Ablations}}{Ablations}}
\label{app:ablation}

\noindent\textbf{Bins.} Without bins (U0~D0), every weight of a group is stored on the middle bin's grid with the flag. On
Llama-2 this is within 0.6\% of {\weshareone{}}, but on Llama-3-8B and the Qwen models it is 8-65\% above it{.
On these models a} few channels per group need the full INT3 grid of the high bin. The menu of {\wesharetwo{}} contains U0~D0, and
{\wesharetwo{}} is below both on every model.

\noindent\textbf{Flag.} Taking bit $b_0$ of the middle and low bins from the parity of the zero point saves the flag, 1/128
bit per weight, but raises perplexity by 5-30\% {across the six models, and} by 13\%, 16\% and 30\% on Llama-2-7B, Llama-2-13B and Llama-3-8B.

\begin{table}[!ht]
\caption{Ablations at 2 bits: WikiText-2 perplexity, calibration on the Pile. INT2 + clip: GPTQ + clip on the INT2 grid, the starting point (Table~\ref{tab:main}). No bins: U0~D0 in every group. Tied $b_0$: U8~D16 with bit $b_0$ taken from the parity of the zero point instead of the flag. \weshareone{}: U8~D16. Equal bits: a layout per group with $\delta=0$ in every group. \wesharetwo{}: a layout per group by Equation~\ref{eq:alloc}. The lowest perplexity of each model is in \textbf{bold}.}
\label{tab:ablation}
\centering
\footnotesize
\setlength{\tabcolsep}{2.6pt}
\begin{tabular}{@{}lc rr r rrr@{}}
\toprule
 & & \multicolumn{2}{c}{Llama-2} & Llama-3 & \multicolumn{3}{c}{Qwen2.5} \\
\cmidrule(lr){3-4}\cmidrule(lr){5-5}\cmidrule(l){6-8}
Method & b/w & 7B & 13B & 8B & 0.5B & 1.5B & 3B \\
\midrule
INT2 + clip & 2.141 & 10.49 & 8.31 & 39.51 & 129.9 & 32.87 & 21.74 \\
No bins & 2.156 & 9.30 & 7.39 & 18.32 & 105.2 & 24.26 & 16.70 \\
Tied $b_0$ & 2.148 & 10.42 & 8.61 & 21.18 & 66.73 & 24.98 & 17.92 \\
\weshareone{} & 2.156 & 9.25 & 7.43 & 16.35 & 63.63 & 20.61 & 15.44 \\
Equal bits & 2.156 & 9.04 & \textbf{7.19} & 17.34 & \textbf{53.46} & 20.98 & \textbf{15.17} \\
\wesharetwo{} & 2.156 & \textbf{8.95} & 7.31 & \textbf{15.19} & 58.02 & \textbf{20.60} & 15.29 \\
\bottomrule
\end{tabular}
\end{table}

\begin{figure}[!ht]
\centering
\includegraphics[width=\columnwidth]{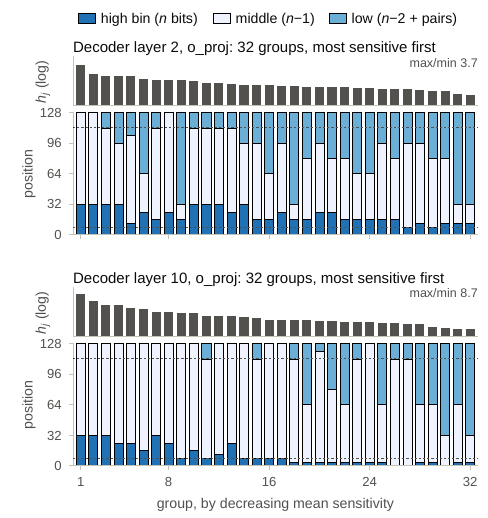}
\caption{{Bin configuration that \wesharetwo{} chooses for the output projection of decoder layers 2 and 10 of
Llama-2-7B at 2 bits (Pile calibration). Its 32 groups of 128 input channels, one attention head each, are sorted by
their mean sensitivity $h_j$ ({grey} strips, log scale){. The} bars show the positions of each group's high, middle and low
bins, as in Figure~\ref{fig:layouts}. Dashed: U8~D16, the layout of \weshareone{}.}}
\label{fig:bins}
\end{figure}

\noindent\textbf{Allocation across groups.} With equal bits in every group, each group can only change the shape of its
layout ($U=D/2$). This lowers perplexity against {\weshareone{}} by 2.6\% on average {but raises it on Llama-3-8B
and Qwen2.5-1.5B, by 6.1\% and 1.8\%. On three models it is below \wesharetwo{}, by up to 7.9\% on Qwen2.5-0.5B.} Moving bits
between groups by Equation~\ref{eq:alloc}{, as \wesharetwo{} does,} lowers it on every model, by 3.6\% on average. The groups
that receive bits are the
more sensitive ones: the rank correlation between a group's mean $h_j$ and its offset is between $+0.38$ and $+0.66$
across the six models. The layouts also differ between families. On Llama-2, where the bins add little, the most frequent
layouts have at most four high-bin positions (on Llama-2-7B, U4~D8 for 17\% of the groups, U4~D0 for 17\% and U0~D0 for
14\%), whereas on the Qwen models the most frequent layout has 16 or 24 (on Qwen2.5-0.5B, U24~D48 for 14\%, U24~D32
for 10\% and U16~D32 for 9\%). {Figure~\ref{fig:bins} shows the layouts that \wesharetwo{} chooses for the output projection of decoder
layers 2 and 10 of Llama-2-7B, where each group is one attention head. The more sensitive groups take larger high bins
and the less sensitive ones larger low bins, with a rank correlation of $+0.70$ and $+0.92$ between a group's mean $h_j$
and its offset. The two layers use 19 and 14 different layouts among their 32 groups, against 8 and 4 in the query
projections of the same layers.}

\section{Hardware Implementation Details}
\label{app:hw}

\subsection{Microarchitecture}
\label{app:hw-arch}

\noindent\textbf{Storage.} Group-rows are stored back to back in row-major order,
without padding, so a group-row may start at any bit. Each record holds the
fields of Table~\ref{tab:budget} in the order the decoder reads them: the scale
$s_{ig}$, the zero point $z_{ig}$ and the flag $f_{ig}$, then the per-weight
fields in position order, then the $D_g/2$ pair bits. A record has
$B_g = (n{-}1)G + \delta(U_g,D_g) + n + 17$ bits (Equation~\ref{eq:rate}).

\noindent\textbf{Layer decoder.} A per-layer table holds one 6-bit layout index per
group, shared by all $R$ rows. A 44-entry ROM maps the index to $(U_g,D_g)$,
and hence to $B_g$; for the one-layout-per-layer variants of Section~\ref{sec:storage} the ROM is
bypassed by a fixed $(U,D)$, with $b_1$ shared by two or four low-bin weights.
A walker requests group-rows in storage order, and a bit aligner cuts each
$B_g$-bit record from a reservoir filled by $W$-bit memory words. Because
every row has the same layout sequence, the walker needs no stored offsets;
for random access, a prefix-sum unit returns the bit offset of every group
within a row. Reserved indices and illegal layouts raise an error and stop
decoding.

\noindent\textbf{Group decoder.} The group decoder (dotted box in
Figure~\ref{fig:decoder}) emits $L$ weights per clock
cycle, at consecutive positions $t = \mathit{base},\dots,\mathit{base}{+}L{-}1$
of one group-row; we call one such output transfer a \emph{beat}.
Following Section~\ref{sec:decoder}, a lane finds the bin of
its position by two comparisons, $t<U_g$ and $t\ge G-D_g$. Its field starts
$n\,h + (n{-}1)\,m + (n{-}2)\,l$ bits into the beat, where $h$, $m$ and $l$
count the high, middle and low positions in $[\mathit{base},t)$. The lane then
forms $q$ by Equation~\ref{eq:codes}, with $f_{ig}$ as $b_0$ for $t\ge U_g$ and the
pair bit $c_{igk}$ as $b_1$ in the low bin. Three registered stages compute
$q-z_{ig}$, the exact integer product of $|q-z_{ig}|$ with the significand of
$s_{ig}$, and a single round-to-nearest-even to binary16, which is Equation~\ref{eq:dequant}
evaluated exactly as for INT-$n$. The decoder also outputs $q-z_{ig}$ for
datapaths that accumulate in integers and apply $s_{ig}$ once per group-row.
Beats may span bin boundaries, so every group-row takes exactly $G/L$ beats irrespective of its layout.

\noindent\textbf{Throughput.} For $G/L\ge 2$, the decoder sustains
$\min(L,\, W G/\bar B)$ weights per cycle, where $\bar B$ is the mean record size: $L$ when the memory
interface keeps up, and the input bandwidth otherwise. With $G=128$ and
$L=16$, a 64-bit interface suffices for every layout of the menu, which needs
at most 54.6 bits per cycle at $n=4$ and 38.5 at $n=3$. In simulation, a row
containing all 44 layouts decodes in exactly one beat per cycle at
$L\in\{4,8,16\}$ for both grids. At $L=32$ with $W=64$, it takes 369 and 544
cycles for 352 beats at $n=3$ and $n=4$.

\noindent\textbf{Permutation.} The decoder emits weights in binned order, so
activations are read through the permutation $\pi$ of Section~\ref{sec:binning}. Since $\pi$ is
shared by all $R$ rows of a layer, an activation vector needs to be permuted
once per token and layer rather than once per row. We therefore report the
decoder without the activation gather.

\subsection{Storage layout}
\label{app:hw-abi}

\begin{itemize}
  \item A layer is stored row by row; each row is the concatenation of its
        $C/G$ group-rows; nothing is padded. The stream is least-significant-bit
        first: bit~0 of the first memory word is bit~0 of the layer.
  \item A group-row is, in this order: $s_{ig}$ (16 bits, IEEE binary16);
        $z_{ig}$ ($n$ bits); $f_{ig}$ (1 bit); the $U_g$, $M_g$ and $D_g$
        fields of $n$, $n{-}1$ and $n{-}2$ bits in position order; and the
        $D_g/S$ shared bits, with $S=2$, or $S=4$ for the four-way variant of
        Section~\ref{sec:storage}. Every field is least-significant-bit first.
  \item Layout index $k\in\{0,\dots,43\}$ enumerates the menu $\Lambda$ in
        lexicographic $(U,D)$ order; indices 44-63 are reserved.
\end{itemize}
The SliM-LLM decoder stores $s$ (16 bits), $z$ ($N{+}1$ bits, wide enough
for the largest precision) and $G$ fields of the group's precision; the INT-$n$
decoder stores $s$, $z$ ($n$ bits), one unused bit and $G$ fields of $n$ bits.

\subsection{Verification}
\label{sec:hw-verif}

We verify the layer decoder against a Python reference model that parses the packed bitstream using a serial bit cursor, independently of the RTL’s per-lane offset calculations. The model reconstructs binary16 weights by rounding the exact product $s_{ig}(q-z_{ig})$ once. Across 62 simulations, the regression checks every output bit and sideband signal, covering:
\begin{itemize}
  \item both quantization grids and lane counts $L\in\{4,8,16,32\}$;
  \item all 44 layouts and random per-group layouts across 86 groups and three rows, matching the input-group count of Llama-2-7B’s down projection;
  \item the layouts U8 with $D\in\{16,48,80,96,120\}$, with two- and four-way sharing, degenerate layouts and group size $G=64$;
  \item memory words of 32, 64 and 128 bits, random input gaps and two consecutive layers.
\end{itemize}

\begin{table*}[!htbp]
\centering
\caption{Logic synthesis per block (Yosys~0.62 and ABC, SkyWater
\texttt{sky130\_fd\_sc\_hd} subset, tt\_025C\_1v80, $G=128$, $L=16$, $W=64$).
Logic delay excludes register timing and interconnect.}
\label{tab:hw-blocks}
\small
\begin{tabular}{@{}lrrrr@{}}
\toprule
Block & Area ($\mu$m$^2$) & Cells & Flip-flops & Logic delay (ns) \\
\midrule
\weshare{} group decoder, $n=4$ & 210{,}243 & 25{,}264 & 2{,}327 & 7.88 \\
\weshare{} group decoder, $n=3$ & 175{,}384 & 20{,}113 & 1{,}957 & 7.63 \\
SliM-LLM group decoder, $\{2,3,4\}$ & 129{,}728 & 13{,}464 & 2{,}169 & 7.36 \\
SliM-LLM group decoder, $\{1,2,3\}$ & 105{,}057 & 11{,}154 & 1{,}799 & 6.13 \\
INT3 group decoder & 100{,}693 & 10{,}095 & 1{,}795 & 6.39 \\
INT2 group decoder & 80{,}106 & 8{,}184 & 1{,}425 & 6.15 \\
\midrule
Bit aligner, records $\le$ 533 bits (\weshare, $n=4$) & 107{,}557 & 15{,}883 & 1{,}184 & 5.03 \\
Bit aligner, $\le$ 532 bits (SliM-LLM, $\{2,3,4\}$) & 107{,}489 & 15{,}817 & 1{,}182 & 4.96 \\
Bit aligner, $\le$ 404 bits (\weshare, $n=3$; INT3) & 82{,}593 & 11{,}413 & 924 & 5.47 \\
Bit aligner, $\le$ 403 bits (SliM-LLM, $\{1,2,3\}$) & 83{,}615 & 11{,}866 & 922 & 4.90 \\
Bit aligner, $\le$ 275 bits (INT2) & 59{,}822 & 8{,}397 & 666 & 4.84 \\
\midrule
Layout ROM (44 entries) & 334 & 66 & 0 & 1.14 \\
Binary16 dequantizer, one lane & 2{,}110 & 352 & 0 & 5.91 \\
\weshare{} layer decoder, $n=4$$^{a}$ & 381{,}838 & 42{,}257 & 5{,}320 & 8.29 \\
\weshare{} layer decoder, $n=3$$^{a}$ & 329{,}181 & 36{,}019 & 4{,}689 & 7.45 \\
\bottomrule
\end{tabular}

\smallskip
\raggedright\footnotesize
$^{a}$Group decoder, aligner, ROM, walker and a 256-entry layout table held in
flip-flops; in a deployment the table belongs in SRAM.
\end{table*}

\subsection{Synthesis results per block}
\label{app:hw-blocks}

Table~\ref{tab:hw-blocks} lists every synthesized block of the hardware implementation: the
group decoders of \weshare{}, SliM-LLM and INT-$n$ at both rates, the bit
aligner of each format, the layout ROM, one binary16 dequantizer lane and
the complete \weshare{} layer decoder. The \weshare{} group decoder takes
1.62$\times$ the area of the SliM-LLM decoder at 3 bits and 1.67$\times$ at
2 bits, with 7-9\% more flip-flops, so most of the extra area is logic. The
bit aligners of \weshare{} and SliM-LLM have almost the same area
(107.5\,k$\mu$m$^2$ at 3 bits), and the layout ROM adds only
334\,$\mu$m$^2$.

\subsection{Weight traffic at equal perplexity}
\label{app:hw-bound}

For batch-1 decoding limited by memory bandwidth, tokens per second are bounded
by the bytes streamed per token. Table~\ref{tab:hw-bound} converts the storage
reductions of Table~\ref{tab:iso} into this bound: the ratio of bytes per token for
SliM-LLM (our runs of its released code) and for \weshare{} at the same
perplexity, counting the quantized linear layers and the FP16 output layer
and excluding the key-value cache. The bounds inherit the assumptions of
Table~\ref{tab:iso}: the one-layout-per-layer configuration of Section~\ref{sec:storage}, at rates
interpolated on its rate-quality curve. They are upper bounds, not measured speedups; since the decoder keeps pace with memory and the per-group decoding work is uniform across rows, a fused kernel can approach them and will be explored in future work.

\subsection{Weight-traffic energy at equal perplexity}
\label{app:hw-energy}

In batch-1 decoding every quantized weight is read from DRAM once per token,
so the energy of weight traffic is $E = N \times b \times e_{\text{bit}}$, where $N$ is the
number of quantized weights, $b$ the bits per weight and $e_{\text{bit}}$ the
DRAM energy per bit transferred. $N$ and $e_{\text{bit}}$ are the same for
every format, so the relative saving is $1-b/b_{\text{ref}}$ and depends on no
assumed constant. 
At comparable storage, \weshare{} uses 2.156 bits per weight against 2.141 for SliM-LLM, approximately 0.7\% more; its benefit at this operating point is improved perplexity. 
Table~\ref{tab:hw-energy} converts the storage saved at 2 bits into weight
energy, with $e_{\text{bit}}\approx 4$\,pJ/bit for HBM2~\cite{oconnor2017fine}
and $\approx 20$\,pJ/bit for off-package DRAM~\cite{horowitz2014energy}.
\weshareone{} reads 13.6-19.4\% fewer weight bits than SliM-LLM at the
same WikiText-2 perplexity (21.6\% or more on C4), which saves
1.2-1.7\,pJ per weight with HBM2 and 5.8-8.3\,pJ per weight with off-package
DRAM. On Llama-2-7B ($6.5\times10^{9}$ quantized weights) this is 336\,MB less
weight traffic per token, or 11-54\,mJ per token. The estimate is
analytical: it covers weight traffic only, not the KV cache, activations or
the FP16 head, which are the same for both formats, and it takes
$e_{\text{bit}}$ from the literature.

\begin{table}[!ht]
\centering
\caption{Upper bound on the decode speedup over SliM-LLM at equal perplexity.}
\label{tab:hw-bound}
\small
\resizebox{\columnwidth}{!}{%
\begin{tabular}{@{}lcccc@{}}
\toprule
 & \multicolumn{2}{c}{3 bits} & \multicolumn{2}{c}{2 bits} \\
\cmidrule(lr){2-3}\cmidrule(lr){4-5}
Model & WikiText-2 & C4 & WikiText-2 & C4 \\
\midrule
Llama-2-7B  & 1.15$\times$ & 1.05$\times$ & 1.19$\times$ & 1.30$\times$ \\
Llama-2-13B & 1.02$\times$ & 1.03$\times$ & 1.20$\times$ & 1.25$\times$ \\
Llama-3-8B  & 1.03$\times$ & 1.08$\times$ & 1.12$\times$ & $\ge$1.23$\times$ \\
\bottomrule
\end{tabular}
}
\end{table}

\begin{table}[!htbp]
  \centering
  \caption{Weight-traffic energy at equal 2-bit WikiText-2 perplexity: bits per weight at which \weshareone{} reaches the perplexity
  of SliM-LLM (2.141 bits per weight), the DRAM energy saved and the saving
  per weight for two DRAM energies.}
  \label{tab:hw-energy}
  \small
  \resizebox{\columnwidth}{!}{%
  \begin{tabular}{lccc}
    \toprule
    Model & Bits per weight & Energy saved & pJ/weight saved \\
          &                 &              & (4 / 20 pJ/bit) \\
    \midrule
    Llama-2-7B  & 1.726 & 19.4\% & 1.66 / 8.31 \\
    Llama-2-13B & 1.745 & 18.5\% & 1.58 / 7.92 \\
    Llama-2-70B & 1.751 & 18.2\% & 1.56 / 7.79 \\
    Llama-3-8B  & 1.758 & 17.9\% & 1.53 / 7.66 \\
    Llama-3-70B & 1.850 & 13.6\% & 1.16 / 5.82 \\
    \bottomrule
  \end{tabular}
  }
\end{table}

\end{document}